\documentclass{article}
\usepackage{arxiv}

\usepackage{sistyle}
\SIthousandsep{,}
\usepackage[noend]{algorithmic}

\renewcommand\algorithmicthen{}

\newcommand{\ONELINEIF}[2]{\STATE \algorithmicif\ #1\ \algorithmicthen\ #2}
\usepackage[utf8]{inputenc}
\usepackage{color}
\usepackage{amssymb}  
\usepackage{hyperref}
\usepackage{ucs}
\usepackage{comment}
\usepackage{amsmath}
\usepackage{pbox}
\usepackage{tcolorbox}
\usepackage{caption}
\usepackage{capt-of}
\usepackage{bm}
\usepackage{graphicx}

\usepackage{array}
\usepackage{multirow}
\usepackage{colortbl}
\usepackage{tikz}
\usepackage{xspace}
\usepackage[caption=false,font=footnotesize]{subfig}

\usepackage{subfig}

\newcommand{\ddpg}{{\sc ddpg}\xspace}

\newcommand{\cmaes}{{\sc cma-es}\xspace}

\newcommand\cem{\textsc{cem}\xspace}

\newcommand{\picma}{{\sc pi$^2$-cma}\xspace}

\newcommand{\rl}{RL\xspace}
\newcommand{\drl}{deep RL\xspace}

\newcommand{\edas}{{\sc EDA}s\xspace}
\newcommand{\eda}{{\sc EDA}\xspace}

\newcommand{\mymath}[1]{\ensuremath{#1}\xspace}
\newcommand{\mymathbf}[1]{\mymath{\mathbf{\boldsymbol{#1}}}}
\newcommand{\covar}{\mymathbf{\Sigma}}

\renewcommand{\vec}[1]{\mymathbf{#1}}

\newcommand\wi[1]{$\circ$}
\newcommand\bu[1]{$\bullet$}
\newcommand\ot[1]{$\star$}
\newcommand\spa[1]{$\spadesuit$}
\newcommand\dn[1]{.}

\newcommand{\params}{{\mymathbf{\theta}}\xspace} 

\newcounter{cbox} 
\newcommand{\cbox}{\arabic{cbox}}

\newcounter{cmes} 
\newcommand{\cmes}{\arabic{cmes}}

\usepackage{algorithmic}
\usepackage{algorithm} 

\makeatletter
\newcounter{algorithmbis}
\renewcommand{\thealgorithmbis}{\thesection.\arabic{algorithmbis}}
\def\algorithmbis{\@ifnextchar[{\@algorithmbisa}{\@algorithmbisb}}
\def\@algorithmbisa[#1]{%
  \refstepcounter{algorithmbis}
  \trivlist
  \leftmargin\z@
  \itemindent\z@
  \labelsep\z@
  \item[\parbox{\linewidth}{%
    \hrule
    \hrule
    \noindent\strut\textbf{Algorithm \thealgorithmbis} #1
    \hrule
  }]\hfil\vskip0em%
}
\def\@algorithmbisb{\@algorithmbisa[]}

\makeatother

\definecolor{myred}{rgb}{0.8,0,0}
\definecolor{mygreen}{rgb}{0,0.6,0}
\definecolor{myblue}{rgb}{0,0,0.7}

\date{}

\newcommand{\vth}{\mymath{\vec \theta}}

\newcommand{\cemrl}{{\sc cem-rl}\xspace}
\newcommand{\cemddpg}{{\sc cem-ddpg}\xspace}
\newcommand{\cemtdd}{{\sc cem-td3}\xspace}
\newcommand{\tdd}{{\sc td3}\xspace}
\newcommand{\relu}{{\sc relu}\xspace}
\newcommand{\evorl}{{\sc erl}\xspace}
\newcommand{\hc}{{\sc half-cheetah-v2}\xspace}
\newcommand{\hop}{{\sc hopper-v2}\xspace}
\newcommand{\sw}{{\sc swimmer-v2}\xspace}
\newcommand{\ant}{{\sc ant-v2}\xspace}
\newcommand{\walk}{{\sc walker2d-v2}\xspace}
\newcommand{\muj}{{\sc mujoco}\xspace}
\newcommand{\gep}{{\sc gep}\xspace}
\newcommand{\geppg}{{\sc gep-pg}\xspace}
\newcommand{\pytorch}{{\sc PyTorch}\xspace}

\newcounter{rules} 
\newcommand{\rules}{\arabic{rules}}

\newcounter{theorems} 
\newcommand{\theorems}{\arabic{theorems}}

\begin{document}

\title{CEM-RL: Combining evolutionary and gradient-based methods for policy search}

\author{Alo\"{i}s Pourchot$^{1,2}$, Olivier Sigaud$^2$\medskip\\
(1) Gleamer\\
  96bis Boulevard Raspail, 75006 Paris, France\\
  {\tt alois.pourchot@gleamer.ai}\medskip\\
(2) Sorbonne Universit\'e, CNRS UMR 7222\\
Institut des Syst\`emes Intelligents et de Robotique, F-75005 Paris, France\\
{\tt olivier.sigaud@upmc.fr}~~~~+33 (0) 1 44 27 88 53}
\maketitle
    \begin{abstract}
Deep neuroevolution and deep reinforcement learning (deep RL) algorithms are two popular approaches to policy search. The former is widely applicable and rather stable, but suffers from low sample efficiency. By contrast, the latter is more sample efficient, but the most sample efficient variants are also rather unstable and highly sensitive to hyper-parameter setting. 
So far, these families of methods have mostly been compared as competing tools. However, an emerging approach consists in combining them so as to get the best of both worlds. Two previously existing combinations use either an ad hoc evolutionary algorithm or a goal exploration process together with the Deep Deterministic Policy Gradient (\ddpg) algorithm, a sample efficient off-policy deep RL algorithm.
In this paper, we propose a different combination scheme using the simple cross-entropy method (\cem) and Twin Delayed Deep Deterministic policy gradient (\tdd), another off-policy deep RL algorithm which improves over \ddpg.
We evaluate the resulting method, \cemrl, on a set of benchmarks classically used in deep RL. We show that \cemrl benefits from several advantages over its competitors and offers a satisfactory trade-off between performance and sample efficiency.
    \end{abstract}

\section{Introduction}

Policy search is the problem of finding a policy or controller maximizing some unknown utility function. 
Recently, research on policy search methods has witnessed a surge of interest due to the combination with deep neural networks, making it possible to find good enough continuous action policies in large domains. From one side, this combination gave rise to the emergence of efficient deep reinforcement learning (deep RL) techniques  \citep{lillicrap2015continuous,schulman2015trust,schulman2017proximal}.
From the other side, evolutionary methods, and particularly deep neuroevolution methods applying evolution strategies (ESs) to the parameters of a deep network have emerged as a competitive alternative to deep RL due to their higher parallelization capability \citep{salimans2016weight,conti2017improving,such2017deep}.

Both families of techniques have clear distinguishing properties. Evolutionary methods are significantly less sample efficient than deep RL methods because they learn from complete episodes, whereas deep RL methods use elementary steps of the system as samples, and thus exploit more information \citep{sigaud2018policy}. In particular, off-policy deep RL algorithms can use a replay buffer to exploit the same samples as many times as useful, greatly improving sample efficiency.
Actually, the sample efficiency of ESs can be improved using the "importance mixing" mechanism, but a recent study has shown that the capacity of importance mixing to improve sample efficiency by a factor of ten is still not enough to compete with off-policy deep RL \citep{pourchot2018importance}.
From the other side, sample efficient off-policy deep RL methods such as the \ddpg algorithm \citep{lillicrap2015continuous} are known to be unstable and highly sensitive to hyper-parameter setting. 
Rather than opposing both families as competing solutions to the policy search problem, a richer perspective consists in combining them so as to get the best of both worlds. As covered in Section~\ref{sec:related}, there are very few attempts in this direction so far. 

After presenting some background in Section~\ref{sec:background}, we propose in Section~\ref{sec:methods} a new combination method that combines the cross-entropy method (\cem) with \ddpg or \tdd, an off-policy deep RL algorithm which improves over \ddpg. In Section~\ref{sec:study}, we investigate experimentally the properties of this \cemrl method, showing its advantages both over the components taken separately and over a competing approach.
Beyond the results of \cemrl, the conclusion of this work is that there is still a lot of unexplored potential in new combinations of evolutionary and deep RL methods.

\section{Related work}
\label{sec:related}

Policy search is an extremely active research domain. The realization that evolutionary methods are an alternative to continuous action reinforcement learning and that both families share some similarity is not new \citep{stulp12icml,stulp2012policy,stulp13paladyn} but so far most works have focused on comparing them \citep{salimans2017evolution,such2017deep,conti2017improving}. Under this perspective, it was shown in \citep{duan2016benchmarking} that, despite its simplicity with respect to most deep RL methods, the Cross-Entropy Method (\cem) was a strong baseline in policy search problems.
Here, we focus on works which combine both families of methods.

Synergies between evolution and reinforcement learning have already been investigated in the context of the so-called {\em Baldwin effect} \citep{simpson1953baldwin}. This literature is somewhat related to research on meta-learning, where one seeks to evolve an initial policy from which a self-learned reinforcement learning algorithm will perform efficient improvement \citep{wang2016learning,houthooft2018evolved,gupta2018meta}.
The key difference with respect to the methods investigated here is that in this literature, the outcome of the RL process is not incorporated back into the genome of the agent, whereas here evolution and reinforcement learning update the same parameters in iterative sequences.

Closer to ours, the work of \cite{colas2018gep} sequentially applies a {\em goal exploration process} (\gep) to fill a replay buffer with purely exploratory trajectories and then applies \ddpg to the resulting data.
The \gep shares many similarities with evolutionary methods, though it focuses on diversity rather than on performance of the learned policies. 
The authors demonstrate on the Continuous Mountain Car and \hc benchmarks that their combination, \geppg, is more sample-efficient than \ddpg, leads to better final solutions and induces less variance during learning. However, due to the sequential nature of the combination, the \gep part does not benefit from the efficient gradient steps of the deep RL part.

Another approach related to ours is the work of \cite{maheswaranathan2018guided}, where the authors introduce optimization problems with a surrogate gradient, i.e. a direction which is correlated with the real gradient. They show that by modifying the covariance matrix of an ES to incorporate the informations contained in the surrogate, a hybrid algorithm can be constructed. They provide a thorough theoretical investigation of their procedure, which they experimentally show capable of outperforming both a standard gradient descent method and a pure ES on several simple benchmarks.
They argue that this method could be useful in \rl, since surrogate gradients appear in Q-learning and actor-critic methods. However, a practical demonstration of those claims remains to be performed. Their approach resembles ours in that they use a gradient method to enhance an ES. But a notable difference is that they use the gradient information to directly change the distribution from which samples are drawn, whereas we use gradient information on the samples themselves, impacting the distribution only indirectly.

The work which is the closest to ours is \cite{khadka2018evolutionaryNIPS}. The authors introduce an algorithm called \evorl (for Evolutionary Reinforcement Learning), which is presented as an efficient combination of a deep \rl algorithm, \ddpg, and a population-based evolutionary algorithm. 
It takes the form of a population of actors, which are constantly mutated and selected in tournaments based on their fitness. In parallel, a single \ddpg agent is trained from the samples generated by the evolutionary population. This single agent is then periodically inserted into the population. When the gradient-based policy improvement mechanism of \ddpg is efficient, this individual outperforms its evolutionary siblings, it gets selected into the next generation and draws the whole population towards higher performance. 
Through their experiments, \citeauthor{khadka2018evolutionaryNIPS} demonstrate that this setup benefits from an efficient transfer of information between the \rl algorithm and the evolutionary algorithm, and vice versa.

However, their combination scheme does not make profit of the search efficiency of ESs. This is unfortunate because ESs are generally efficient evolutionary methods, and importance mixing can only be applied in their context to bring further sample efficiency improvement.

By contrast with the works outlined above, the method presented here combines \cem and \tdd in such a way that our algorithm benefits from the gradient-based policy improvement mechanism of \tdd, from the better stability of ESs, and may even benefit from the better sample efficiency brought by importance mixing, as described in Appendix~\ref{sec:mixing}.

\section{Background}
\label{sec:background}

In this section, we provide a quick overview of the evolutionary and deep RL methods used throughout the paper.

\subsection{Evolutionary algorithms, evolution strategies and EDAs}
\label{sec:eas}

Evolutionary algorithms manage a limited population of individuals, and generate new individuals randomly in the vicinity of the previous {\em elite} individuals. There are many variants of such algorithms, some using tournament selection as in \cite{khadka2018evolutionaryNIPS}, niche-based selection or more simply taking a fraction of elite individuals, see \cite{back1996evolutionary} for a broader view.
Evolution strategies can be seen as specific evolutionary algorithms where only one individual is retained from one generation to the next, this individual being the mean of the distribution from which new individuals are drawn. More specifically, an optimum individual is computed from the previous samples and the next samples are obtained by adding Gaussian noise to the current optimum.
Finally, among ESs, Estimation of Distribution Algorithms (\edas) are a specific family where the population is represented as a distribution using a covariance matrix \covar \citep{larranaga2001estimation}.
This covariance matrix defines a multivariate Gaussian function and samples at the next iteration are drawn according to \covar.
Along iterations, the ellipsoid defined by $\covar$ is progressively adjusted to the top part of the hill corresponding to the local optimum $\params^*$.
Various instances of \edas, such as the  Cross-Entropy Method (\cem), Covariance Matrix Adaptation Evolution Strategy (\cmaes) and \picma, are covered in \cite{stulp12icml,stulp2012policy,stulp13paladyn}. Here we focus on the first two.

\subsection{The Cross-Entropy Method and \cmaes}
\label{sec:cem_cmaes}

The Cross-Entropy Method (\cem) is a simple \eda where the number of elite individuals is fixed to a certain value $K_e$ (usually set to half the population). After all individuals of a population are evaluated, the $K_e$ fittest individuals are used to compute the new mean and variance of the population, from which the next generation is sampled after adding some extra variance $\epsilon$ to prevent premature convergence.

In more details, each individual $x_i$ is sampled by adding Gaussian noise around the mean of the distribution $\mu$, according to the current covariance matrix $\Sigma$, i.e.\ $x_i \sim \mathcal{N}(\mu, \Sigma)$. The problem-dependent fitness of these new individuals $(f_i)_{i=1,\dots,\lambda}$ is computed, and the top-performing $K_e$ individuals, $(z_i)_{i=1,\dots,K_e}$ are used to update the parameters of the distribution as follows:

\begin{align}
\label{eq:cem}
\mu _{new} &= \sum_{i=1}^{K_e}\lambda_i z_i \\
\label{eq:cem2}
\Sigma_{new} &= \sum_{i=1}^{K_e}\lambda_i(z_i-\mu_{old})(z_i-\mu_{old})^T + \epsilon \mathcal{I},
\end{align}

where $(\lambda_i)_{i=1,\dots,K_e}$ are weights given to the individuals, commonly chosen with $\lambda_i = \frac{1}{K_e}$ or $\lambda_i = \frac{log(1 + K_e) / i}{\sum_{i=1}^{K_e}\log(1 + K_e) / i}$ \citep{hansen2016cma}. In the former, each individual is given the same importance, whereas the latter gives more importance to better individuals.

Similarly to \cem, Covariance Matrix Adaptation Evolution Strategy (\cmaes) is an \eda where the number of elite individuals is fixed to a certain value $K_e$. The mean and covariance of the new generation are constructed from those individuals. However this construction is more elaborate than in \cem. The top $K_e$ individuals are ranked according to their performance, and are assigned weights based on this ranking. Those weights in turn impact the construction of the new mean and covariance. Quantities called "Evolutionary paths" are also used to accumulate the search directions of successive generations. In fact, the updates in \cmaes are shown to approximate the natural gradient, without explicitly modeling the Fisher information matrix \citep{arnold11informationgeometric}.

A minor difference between \cem and \cmaes can be found in the update of the covariance matrix. In its standard formulation, \cem uses the new estimate of the mean $\mu$ to compute the new $\Sigma$, whereas \cmaes uses the current $\mu$ (the one that was used to sample the current generation) as is the case in \eqref{eq:cem2}. We used the latter as \cite{hansen2016cma} shows it to be more efficient. The algorithm we are using can thus be described either as \cem using the current $\mu$ for the estimation of the new $\Sigma$, or as \cmaes without evolutionary paths. The difference being minor, we still call the resulting algorithm \cem. Besides,  we add some noise in the form of $\epsilon \mathcal{I}$ to the usual covariance update to prevent premature convergence. We choose to have an exponentially decaying $\epsilon$, by setting an initial and a final standard deviation, respectively $\sigma_{init}$ and $\sigma_{end}$, initializing $\epsilon$ to $\sigma_{init}$
and updating $\epsilon$ at each iteration with 
$\epsilon = \tau_{cem}\epsilon + (1 - \tau_{cem})\sigma_{end}$.

Note that, in practice \covar can be too large for computing the updates and sampling new individuals. Indeed, if $n$ denotes the number of actor parameters, simply sampling from \covar scales at least in $\mathcal{O}(n^{2.3})$, which becomes quickly intractable. Instead, we constrain \covar to be diagonal. This means that in our computations, we replace the update in \eqref{eq:cem2} by

\begin{equation}
\covar_{new} = \sum_{i=1}^{K_e}\lambda_i(z_i-\mu_{old})^2 + \epsilon \mathcal{I},
\end{equation}
where the square of the vectors denote the vectors of the square of the coordinates.

\subsection{DDPG and TD3}
\label{sec:deepRL}

The Deep Deterministic Policy Gradient (\ddpg) \citep{lillicrap2015continuous} and Twin Delayed Deep Deterministic policy gradient (\tdd) \citep{fujimoto2018adressing} algorithms are two off-policy, actor-critic and sample efficient deep RL algorithms. The \ddpg algorithm suffers from instabilities partly due to an overestimation bias in  critic updates, and is known to be difficult to tune given its sensitivity to hyper-parameter settings. The availability of properly tuned code baselines incorporating several advanced mechanisms improves on the latter issue \citep{baselines}. The \tdd algorithm rather improves on the former issue, limiting the over-estimation bias by using two critics and taking the lowest estimate of the action values in the update mechanisms. 

\section{Methods}
\label{sec:methods}

\begin{figure}[!ht]
  \centering
\subfloat[\label{fig:cemrl_architecture}]{\includegraphics[width=0.48\linewidth]{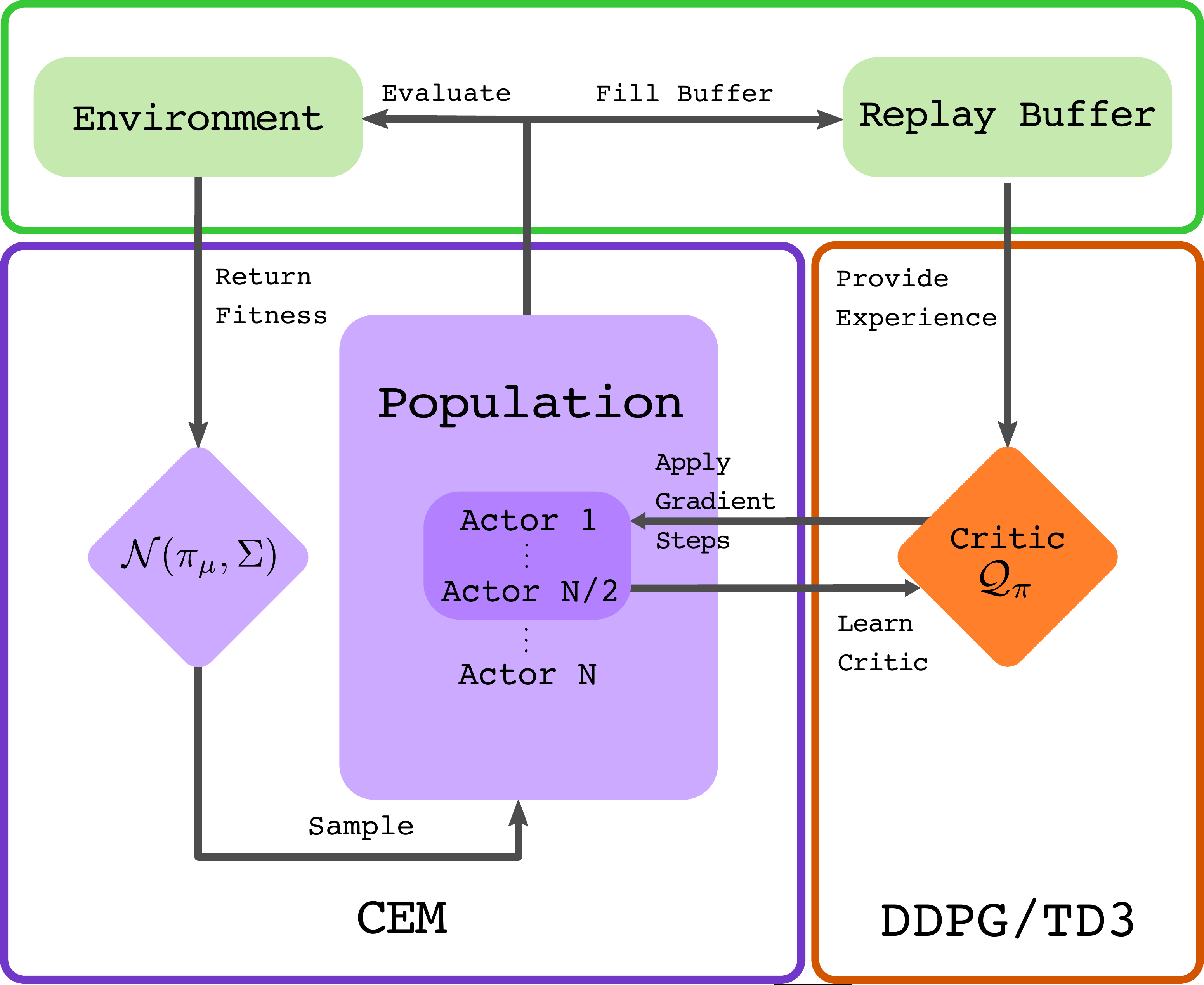}} \hfill
  \subfloat[\label{fig:erl_architecture}]{\includegraphics[width=0.48\linewidth]{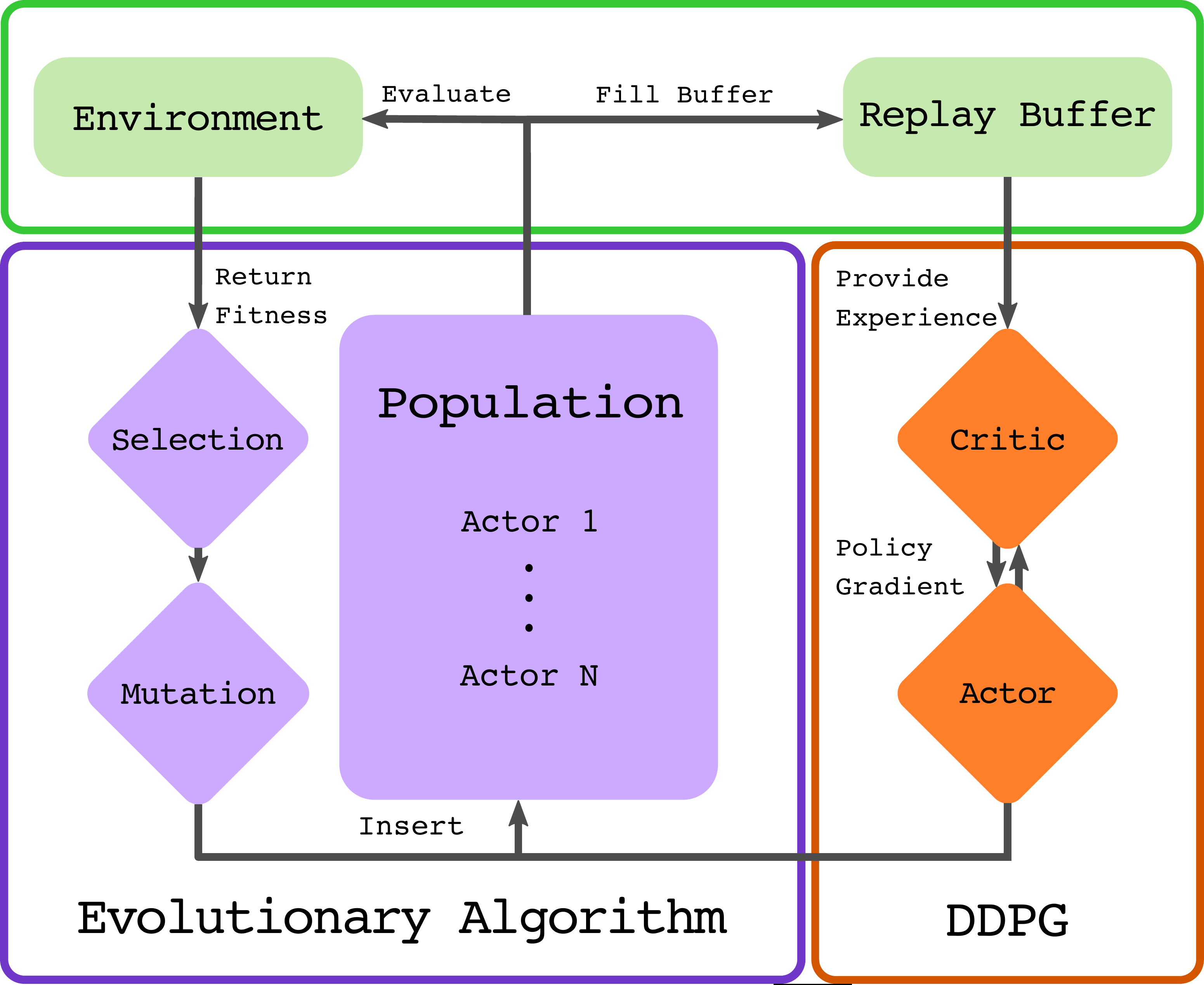}} \hfill
   \caption{Architectures of the \cemrl (a)  and \evorl (b) algorithms \label{fig:archi}}
\end{figure}

As shown in \figurename~\ref{fig:cemrl_architecture}, the \cemrl method combines \cem with either \ddpg or \tdd, giving rise to two algorithms named \cemddpg and \cemtdd. The mean actor of the \cem population, referred to as $\pi_\mu$, is first initialized with a random actor network. A unique critic network $\mathcal{Q^\pi}$ managed by \tdd or \ddpg is also initialized. At each iteration, a population of actors is sampled by adding Gaussian noise around the current mean $\pi_\mu$, according to the current covariance matrix \covar. Half of the resulting actors are directly evaluated. The corresponding fitness is computed as the cumulative reward obtained during an episode in the environment.
Then, for each actor of the other half of the population, the critic is updated using this actor and, reciprocally, the actor follows the direction of the gradient given by the critic $\mathcal{Q^\pi}$ for a fixed number of steps. The resulting actors are evaluated after this process. The \cem algorithm then takes the top-performing half of the resulting global population to compute its new $\pi_\mu$ and \covar.
The steps performed in the environment used to evaluate all actors in the population are fed into the replay buffer. The critic is trained from that buffer pro rata to the quantity of new information introduced in the buffer at the current generation. For instance, if the population contains 10 individuals, and if each episode lasts 1000 time steps, then \num{10000} new samples are introduced in the replay buffer at each generation.  The critic is thus trained for \num{10000} mini-batches, which are divided into 2000 mini-batches per learning actor. This is a common practice in deep \rl algorithms, where one mini-batch update is performed for each step of the actor in the environment. We also choose this number of steps (\num{10000}) to be the number of gradient steps taken by half of the population at the next iteration.
A pseudo-code of \cemrl is provided in Algorithm~\ref{alg:cemrl}.

\begin{algorithm}[htb]
  \caption{\cemrl}
  \label{alg:cemrl}
  \begin{algorithmic}[1]
  \REQUIRE \textit{max\_steps}, the maximum number of steps in the environment \\
  \hspace{22pt} $\tau_\cem, \sigma_{init}, \sigma_{end}$ and \textit{pop\_size}, hyper-parameters of the \cem algorithm \\
  \hspace{22pt} $\gamma, \tau, lr_{actor}$ and $lr_{critic}$, hyper-parameters of \ddpg
  \item[]
  \STATE Initialize a random actor $\pi_\mu$, to be the mean of the \cem algorithm
  \STATE Let $\Sigma = \sigma_{init} \mathcal{I}$ be the covariance matrix of the \cem algorithm 
  \STATE Initialize the critic $\mathcal{Q}^\pi$ and the target critic $\mathcal{Q}_t^\pi$ 
  \STATE Initialize an empty cyclic replay buffer $\mathcal{R}$ 
  \item[]
  \STATE \textit{total\_steps, actor\_steps} = $0, 0$
  \WHILE{\textit{total\_steps} $<$ \textit{max\_steps}:}
        \item[]
        \STATE Draw the current population \textit{pop} from $\mathcal{N}(\pi_\mu, \Sigma)$ with importance mixing (see Algorithm~\ref{alg:IM} in Appendix~\ref{sec:mixing})
        \FOR{$i \leftarrow 1$ to $pop\_size / 2$:}
        	\STATE Set the current policy $\pi$ to $pop[i]$
        	\STATE Initialize a target actor $\pi_t$ with the weights of $\pi$
            \STATE Train $\mathcal{Q}^\pi$ for 2 $*$ \textit{actor\_steps} / \textit{pop\_size} mini-batches
            \STATE Train $\pi$ for \textit{actor\_steps} mini-batches
            \STATE Reintroduce the weights of $\pi$ in \textit{pop}
        \ENDFOR
        \item[]
        \STATE \textit{actor\_steps} = $0$
        \FOR{$i \leftarrow 1$ to $pop\_size$:}
        	\STATE Set the current policy $\pi$ to $pop[i]$
            \STATE (fitness $f$, steps $s$) $\leftarrow$ evaluate($\pi$) 
            \STATE Fill $\mathcal{R}$ with the collected experiences
            \STATE \textit{actor\_steps} = \textit{actor\_steps} + $s$
        \ENDFOR
        \textit{total\_steps} = \textit{total\_steps} + \textit{actor\_steps}
        
        \item[]
        \STATE Update $\pi_\mu$ and \covar with the top half of the population (see \eqref{eq:cem} and \eqref{eq:cem2} in Section~\ref{sec:cem_cmaes})
        
	\ENDWHILE
    \item[]
  \STATE \textbf{end while}
  \end{algorithmic}
\end{algorithm}

In cases where applying the gradient increases the performance of the actor, \cem benefits from this increase by incorporating the corresponding actors in its computations. By contrast, in cases where the gradient steps decrease performance, the resulting actors are ignored by \cem, which instead focuses on standard samples around $\pi_\mu$. Those poor samples do not bring new insight on the current distribution of the \cem algorithm, since the gradient steps takes them away from the current distribution. However, since all evaluated actors are filling the replay buffer, the resulting experience is still fed to the critic and the future learning actors, providing some supplementary exploration.

This approach generates a beneficial flow of information between the deep \rl part and the evolutionary part. Indeed, on one hand, good actors found by following the current critic directly improve the evolutionary population. On the other hand, good actors found through evolution fill the replay buffer from which the \rl algorithm learns.

In that respect, our approach benefits from the same properties as the \evorl algorithm \citep{khadka2018evolutionaryNIPS} depicted in \figurename~\ref{fig:erl_architecture}.
But, by contrast with \cite{khadka2018evolutionaryNIPS}, gradient steps are directly applied to several samples, and using the \cem algorithm makes it possible to use importance mixing, as described in Appendix~\ref{sec:mixing}.
Another difference is that in \cemrl gradient steps are applied at each iteration whereas in \evorl, a deep \rl actor is only injected to the population from time to time.
One can also see from \figurename~\ref{fig:archi} that, in contrast to \evorl, \cemrl does not use any deep \rl actor.
Other distinguishing properties between \evorl and \cemrl are discussed in the light of empirical results in Section~\ref{sec:results}.

Finally, given that \cmaes is generally considered as more sophisticated than \cem, one may wonder why we did not use \cmaes instead of \cem into the \cemrl algorithm. Actually, the key contribution of \cmaes with respect to \cem consists of the evolutionary path mechanism (see Section~\ref{sec:cem_cmaes}), but this mechanism results in some inertia in \covar updates, which resists to the beneficial effect of applying RL gradient steps.

\section{Experimental study}
\label{sec:study}

In this section, we study the \cemrl algorithm to answer the following questions: 
\begin{itemize}
\item 
How does the performance of \cemrl compare to that of \cem and \tdd taken separately? What if we remove the \cem mechanism, resulting in a multi-actor \tdd?
\item 
How does \cemrl perform compared to \evorl?
What are the main factors explaining the difference between both algorithms?
\end{itemize}

Additionally, in Appendices~\ref{sec:mixing} to \ref{sec:sw}, we investigate other aspects of the performance of \cemrl such as the impact of importance mixing, the addition of action noise or the use of the $tanh$ non-linearity.

\subsection{Experimental setup}
\label{sec:setup}
In order to investigate the above questions, we evaluate the corresponding algorithms in several continuous control tasks simulated with the \muj physics engine and commonly used as policy search benchmarks: \hc, \hop,  \walk, \sw and \ant \citep{brockman2016openai}.

We implemented \cemrl with the \pytorch library \footnote{The code for reproducing the 
experiments is available at {\tt https://github.com/apourchot/CEM-RL}.}. We built our code around the \ddpg and \tdd implementations given by the authors of the \tdd algorithm\footnote{Available at {\tt https://github.com/sfujim/TD3}.}. For the \evorl implementation, we used one given by the authors\footnote{Available at {\tt https://github.com/ShawK91/erl\_paper\_nips18}.}. 

Unless specified otherwise, each curve represents the average over 10 runs of the corresponding quantity, and the variance corresponds to the $68\%$ confidence interval for the estimation of the mean. In all learning performance figures, dotted curves represent medians and the x-axis represents the total number of steps actually performed in the environment, to highlight potential sample efficiency effects, particularly when using importance mixing (see Appendix~\ref{sec:mixing}). 

Architectures of the networks are described in Appendix~\ref{sec:archi}. Most \tdd and \ddpg hyper-parameters were reused from \cite{fujimoto2018adressing}. The only notable difference is the use of $tanh$ non linearities instead of \relu in the actor network, after we spotted that the latter performs better than the former on several environments. We trained the networks with the Adam optimizer \citep{kingma2014adam}, with a learning rate of $1e^{-3}$ for both the actor and the critic. The discount rate $\gamma$ was set to 0.99, and the target weight $\tau$ to $5e^{-3}$. All populations contained 10 actors, and the standard deviations $\sigma_{init}$, $\sigma_{end}$ and the constant $\tau_{cem}$ of the \cem algorithm were respectively set to $1e^{-3}$, $1e^{-5}$ and 0.95. Finally, the size of the replay buffer was set to $1e^6$, and the batch size to $100$.

\subsection{Results}
\label{sec:results}

We first compare \cemtdd to \tdd, \tdd and a multi-actor variant of \tdd, then \cemrl to \evorl based on several benchmarks. A third section is devoted to additional results which have been rejected in appendices to comply with space constraints.

\subsubsection{Comparison to \cem, \tdd and a multi-actor \tdd}
\label{sec:res1}

\begin{figure}[!ht]
  \centering
  \subfloat[\label{fig:hc_comparison}]{\includegraphics[width=0.34\linewidth]{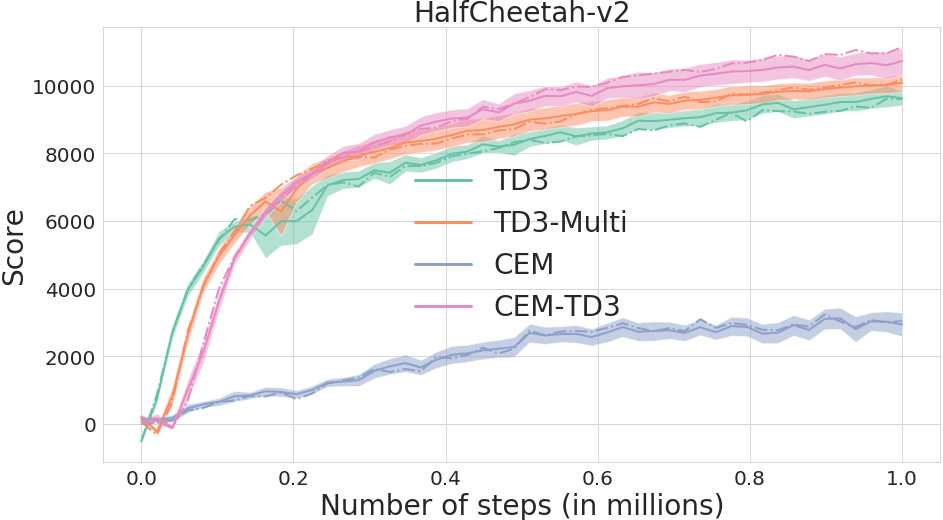}}
  \subfloat[\label{fig:hopper_comparison}]{\includegraphics[width=0.34\linewidth]{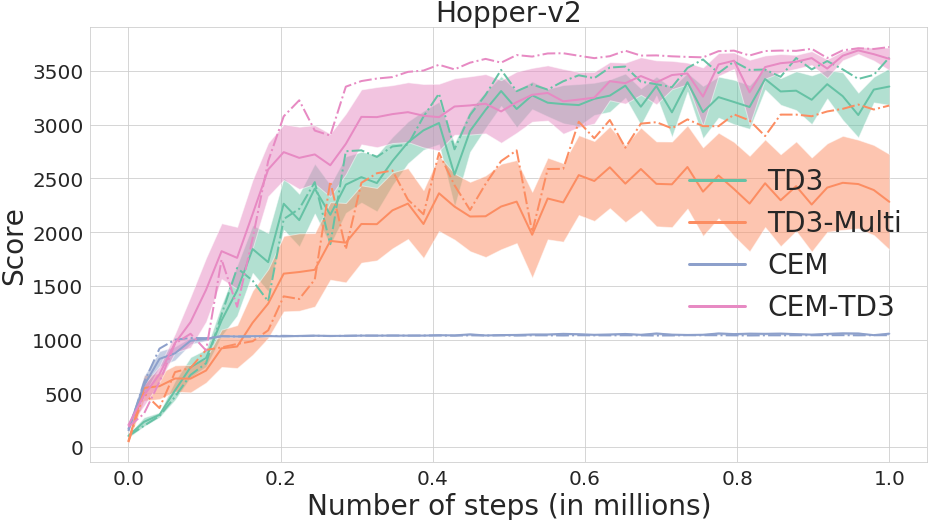}}
  \subfloat[\label{fig:walker_comparison}]{\includegraphics[width=0.34\linewidth]{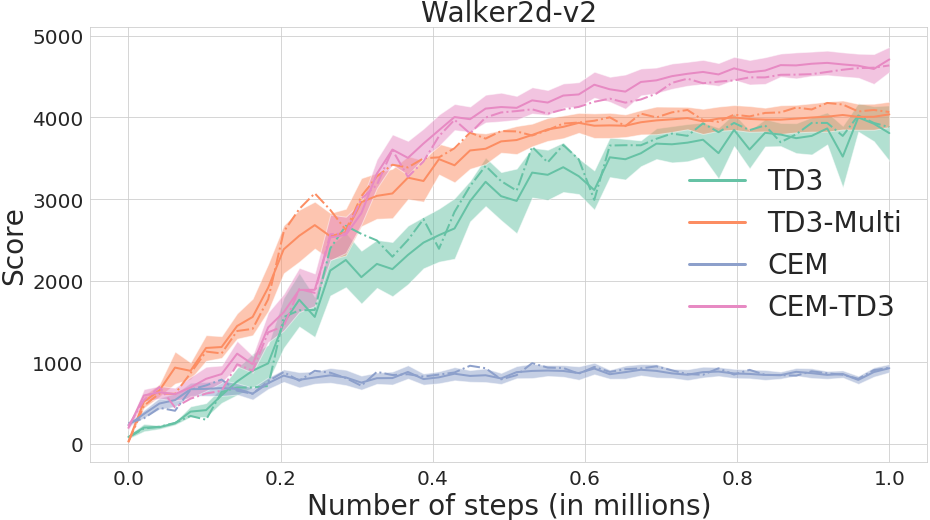}}
\caption{Learning curves of \tdd, \cem and \cemrl on the \hc, \hop, and \walk benchmarks. 
\label{fig:cem_tdd_cmerl}}
\end{figure}

In this section, we compare \cemtdd to three baselines: our variant of \cem, \tdd and a multi-actor variant of \tdd. For \tdd and its multi-actor variant, we report the average of the score of the agent over 10 episodes for every \num{5000} steps performed in the environment. For \cem and \cemtdd, we report after each generation the average of the score of the new average individual of the population over 10 episodes. 
From \figurename~\ref{fig:cem_tdd_cmerl}, one can see that \cemtdd outperforms \cem and \tdd on \hc, \hop and \walk. On most benchmarks, \cemtdd also displays slightly less variance than \tdd. Further results in Appendix~\ref{app:ant} show that on \ant, \cemtdd outperforms \cem and is on par with \tdd. More surprisingly, \cem outperforms all other algorithms on \sw, as covered in Appendix~\ref{sec:sw}.

\begin{table}[ht]
{\small
\centering
\begin{tabular}{l r r r r r r }

\hline 
         & \multicolumn{3}{c}{\cem} & \multicolumn{3}{c}{\tdd}  \\
         Environment & Mean & Var. & Median  & Mean & Var. & Median  \\
         \hline
         \\
  \hc & $2940$ & $12\%$ & $3045$ & $9630$ & $2.1\%$ & $9606$ \\
  
  \hop &  $1055$ & $1.3\%$ & $1040$ & $3355$ & $5.1\%$ & $3626$ \\
  
 \walk & $928$ & $5.4\%$ & $934$ & $3808$ & $8.9\%$ & $3882$ \\
 
  \sw & $\mathbf{351}$ & $\mathbf{2.7\%}$ & $\mathbf{361}$ & $63$ & $14\%$ & $47$ \\
  
  \ant & $487$ & $6.7\%$ & $506$ & $4027$ & $10\%$ & $4587$  \\
  
  \\
  \hline
\hline 
         & \multicolumn{3}{c}{\tdd Multi-Actor}  & \multicolumn{3}{c}{\cemtdd} \\
         Environment & Mean & Var. & Median  & Mean & Var. & Median \\
         \hline
         \\
  \hc & $9662$ & $2.8\%$ & $9710$ & $\mathbf{10725}$ & $\mathbf{3.7\%}$ & $\mathbf{11539}$ \\
  
  \hop &  $2056$ & $20\%$ & $2376$ & $\mathbf{3613}$ & $\mathbf{2.9\%}$ & $\mathbf{3722}$ \\
   
  \walk & $3934$ & $4.1\%$ & $3954$ & $\mathbf{4711}$ & $\mathbf{3.3\%}$ & $\mathbf{4637}$ \\
  
  \sw & $76$ & $14\%$ & $60$ & $75$ & $15\%$ & $62$ \\
  
  \ant & $3567$ & $22\%$ & $3911$ & $\mathbf{4251}$ & $\mathbf{5.9\%}$ & $\mathbf{4310}$ \\
 
  \\
  \hline
  \hline
  \end{tabular}
\caption{\label{table:summary} Final performance of \cem, \tdd, multi-actor \tdd and \cemtdd on 5 environments. We report the mean ands medians over 10 runs of 1 million steps. 
For each benchmark, we highlight the results of the method with the best mean.}
}
\end{table}

One may wonder whether the good performance of \cemtdd mostly comes from its "ensemble method" nature \citep{osband2016deep}. Indeed, having a population of actors improves exploration and stabilizes performances by filtering out instabilities that can appear during learning. 
To answer this question, we performed an ablative study where we removed the \cem mechanism.
We considered a population of 5 actors initialized as in \cemtdd, but then just following the gradient given by the \tdd critic. This algorithm can be seen as a multi-actor \tdd where all actors share the same critic. We reused the hyper-parameters described in Section~\ref{sec:results}. From \figurename~\ref{fig:cem_tdd_cmerl}, one can see that \cemtdd outperforms more or less significantly multi-actor \tdd on all benchmarks, which clearly suggests that the evolutionary part contributes to the performance of \cemtdd.


As a summary, Table~\ref{table:summary} gives the final performance of methods compared in this Section. We conclude that \cemtdd is generally superior to \cem, \tdd and multi-actor \tdd. More precisely, in environments where \tdd provides a useful gradient information, \cemtdd enhances \cem by accelerating updates towards better actors, and it enhances \tdd by reducing variance in the learning process. 

\subsubsection{Comparison to \evorl}
\label{sec:res2}

\begin{figure}[!ht]
  \centering
  \subfloat[\label{fig:erl_cemrl_hc}]{\includegraphics[width=0.34\linewidth]{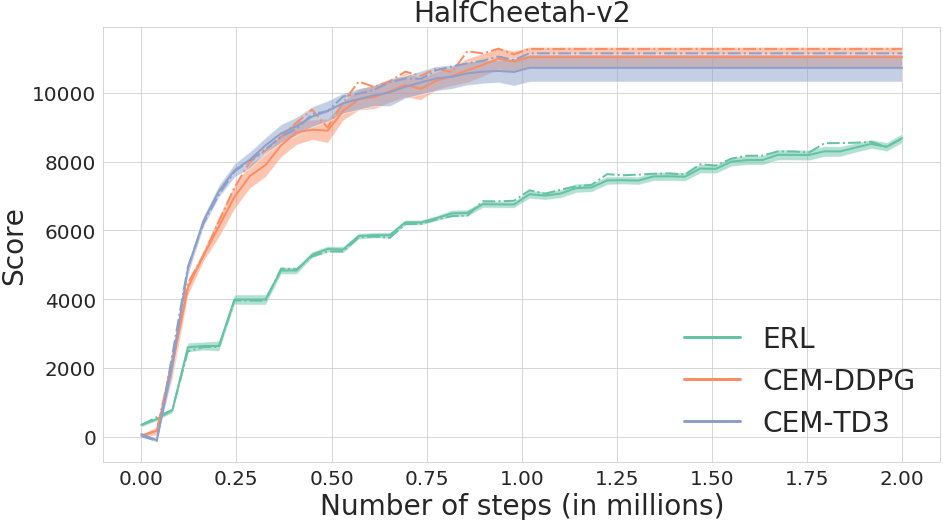}}
  \subfloat[\label{fig:erl_cemrl_hopper}]{\includegraphics[width=0.34\linewidth]{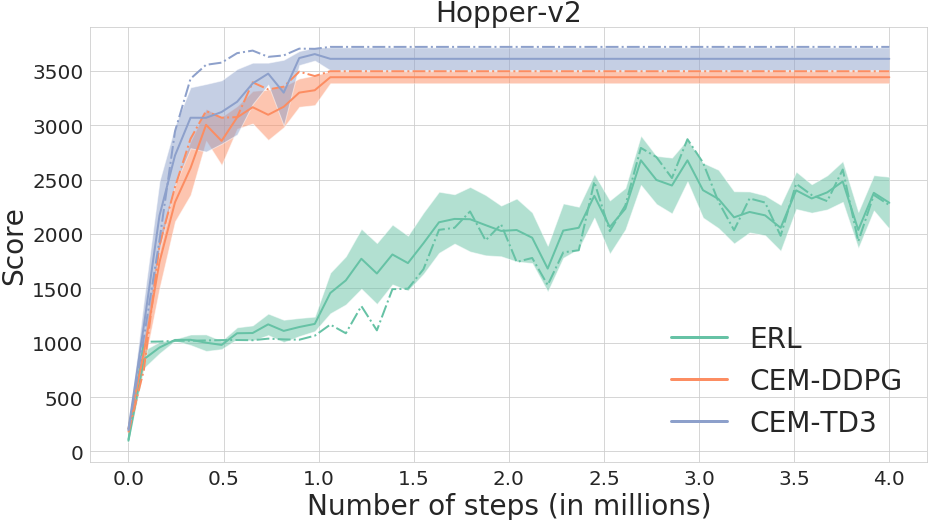}}
  \subfloat[\label{fig:erl_cemrl_walker}]{\includegraphics[width=0.34\linewidth]{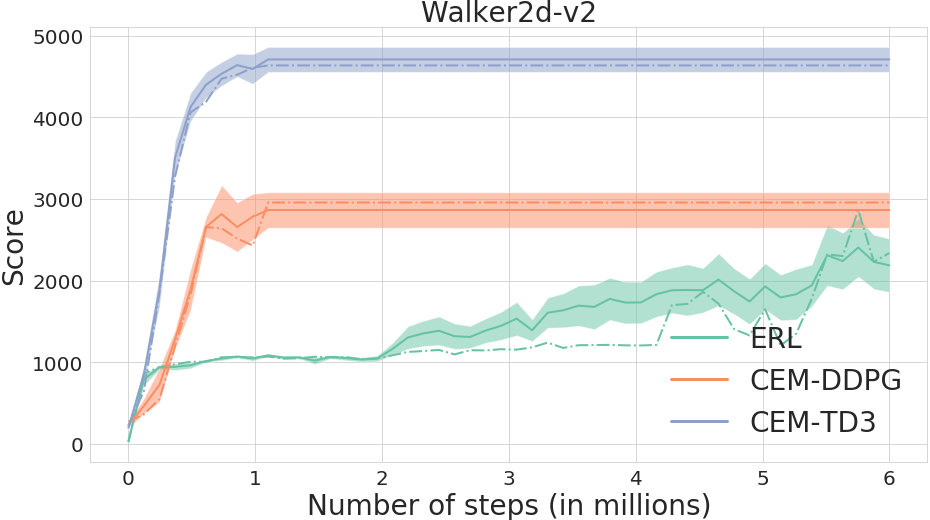}}
   \caption{Learning curves of \evorl, \cemddpg and \cemtdd on \hc, \hop, \ant and \walk. Both \cemrl methods are only trained 1 million steps. 
   \label{fig:erl_cemrl}}
\end{figure}

In this section, we compare \cemrl to \evorl. The \evorl method using \ddpg rather than \tdd, we compare it to both \cemddpg and \cemtdd. This makes it possible to isolate the effect of the combination scheme from the improvement brought by \tdd itself. Results are shown in \figurename~\ref{fig:erl_cemrl}.   
We let \evorl learn for the same number of steps as in \citeauthor{khadka2018evolutionaryNIPS}, namely 2 millions on \hc and \sw, 4 millions on \hop, 6 millions on \ant and 10 millions on \walk. However, due to limited computational resources, we stop learning with both \cemrl methods after 1 million steps, hence the constant performance after 1 million steps.

Our results slightly differ from those of the \evorl paper \citep{khadka2018evolutionaryNIPS}. We explain this difference by two factors. First, the authors only average their results over 5 different seeds, whereas we used 10 seeds. Second, the released implementation of \evorl may be slightly different from the one used to produce the published results\footnote{personal communication with the authors}, raising again the reproducibility issue recently discussed in the reinforcement learning literature \citep{henderson2017deep}.

\begin{table}[ht]
\centering
{\small
\scalebox{0.95}{%
\begin{tabular}{l  r r r  r r r  r r r }
\hline 
         & \multicolumn{3}{c}{\evorl}  & \multicolumn{3}{c}{\cemddpg} & \multicolumn{3}{c}{\cemtdd} \\
         Environment & Mean & Var. & Median  & Mean & Var. & Median & Mean & Var. & Median \\
         \hline
         &&&&&&&&&\\
  \hc & $8684$ & $1.5\%$ & $8675$ & $\mathbf{11035}$ & $\mathbf{2.7\%}$ & $\mathbf{11276}$ & $10725$ & $3.7\%$ & $11539$\\
  
  \hop &  $2288$ & $10.5\%$ & $2267$ & $3444$ & $1.6\%$ & $3499$ & $\mathbf{3613}$ & $\mathbf{2.9\%}$ & $\mathbf{3722}$\\
  
  \walk & $2188$ & $15\%$ & $2338$ & $2865$ & $7.6\%$ & $2958$& $\mathbf{4711}$ & $\mathbf{3.3\%}$ & $\mathbf{4637}$ \\
  
  \sw & $\mathbf{350}$ & $\mathbf{2.41\%}$ & $\mathbf{360}$ & $268$ & $12\%$ & $279$& $75$ & $15\%$ & $62$ \\
  
  \ant & $3716$ & $18.1\%$ & $4240$ & $2170$ & $52\%$ & $3574$ & $\mathbf{4251}$ & $\mathbf{5.9\%}$ & $\mathbf{4310}$\\
  
  &&&&&&&&&\\
  \hline
  \hline
  \end{tabular}
  }
  }
\caption{\label{table:ddpg_tdd} Final performance of \evorl, \cemddpg and \cemtdd on 5 environments. We report the mean ands medians over 10 runs of 1 million steps. 
For each benchmark, we highlight the results of the method with the best mean.}
\end{table}

\figurename~\ref{fig:erl_cemrl} shows that after performing 1 million steps, both \cemrl methods outperform \evorl on \hc, \hop and \walk. We can also see that \cemtdd outperforms \cemddpg on \walk. 
On \ant, \cemddpg and \evorl being on par after 1 million steps, we increased the number of learning steps in \cemddpg to 6 millions. The corresponding results are shown in \figurename~\ref{fig:erl_cemrl_ant} in Appendix~\ref{app:ant}. Results on \sw are covered in Appendix~\ref{sec:sw}. 

One can see that, beyond outperforming \evorl, \cemtdd outperforms \cemddpg on most benchmarks, in terms of final performance, convergence speed, and learning stability. This is especially true for hard environments such as \walk and \ant. The only exception in \sw, as studied in Appendix~\ref{sec:sw}.

Table~\ref{table:ddpg_tdd} gives the final best results of methods used in this Section. The overall conclusion is that \cemrl generally outperforms \evorl.

\subsubsection{Additional results}
\label{sec:other_res}

In this section, we outline the main messages arising from further studies that have been rejected in appendices in order to comply with space constraints.

In Appendix~\ref{sec:mixing}, we investigate the influence of the importance mixing mechanism over the evolution of performance, for \cem and \cemrl. Results show that importance mixing has a limited impact on the sample efficiency of \cemtdd on the benchmarks studied here, in contradiction with results from \cite{pourchot2018importance} obtained using various standard evolutionary strategies. The fact that the covariance matrix \covar moves faster with \cemrl may explain this result, as it prevents the reuse of samples.

In Appendix~\ref{sec:noise}, we analyze the effect of adding Gaussian noise to the actions of \cemtdd. Unlike what \cite{khadka2018evolutionaryNIPS} suggested using \evorl, we did not find any conclusive evidence that action space noise improves performance with \cemtdd. This may be due to the fact that, as further studied in Appendix~\ref{sec:dynam},
the evolutionary algorithm in \evorl tends to converge to a unique individual, hence additional noise is welcome, whereas evolutionary strategies like \cem more easily maintain some exploration.
Indeed, we further investigate the different dynamics of parameter space exploration provided by the \evorl and \cemtdd algorithms in Appendix~\ref{sec:dynam}. \figurename~\ref{fig:dynam} and \ref{fig:erl_hist} show that the evolutionary population in \evorl tends to collapse towards a single individual, which does not happen with the \cem population due to the sampling method.

In Appendix~\ref{sec:sw}, we highlight the fact that, on the \sw benchmark, the performance of the algorithms studied in this paper varies a lot from the performance obtained on other benchmarks. The most likely explanation is that, in \sw, any deep RL method provides a deceptive gradient information which is detrimental to convergence towards efficient actor parameters. In this particular context, \evorl better resists to detrimental gradients than \cemrl, which suggests to design a version of \evorl using \cem to improve the population instead of its ad hoc evolutionary algorithm.

Finally, in Appendix~\ref{sec:relu}, we show that using a $tanh$ non-linearity in the architecture of actors often results in significantly stronger performance than using \relu. This strongly suggests performing "neural architecture search" \citep{zoph2016neural,elsken2018nassurvey} in the context of \rl.

  \section{Conclusion and future work}
  \label{sec:conclu}

We advocated in this paper for combining evolutionary and deep RL methods rather than opposing them. In particular, we have proposed such a combination, the \cemrl method, and showed that in most cases it was outperforming not only some evolution strategies and some sample efficient off-policy deep RL algorithms, but also another combination, the \evorl algorithm. Importantly, despite being mainly an evolutionary method, \cemrl is competitive to the state-of-the-art even when considering sample efficiency, which is not the case of other deep neuroevolution methods
\citep{salimans2017evolution,such2017deep}.

Beyond these positive performance results, our study raises more fundamental questions. First, why does the simple \cem algorithm perform so well on the \sw benchmark? Then, our empirical study of importance mixing did not confirm a clear benefit of using it, neither did the effect of adding noise on actions.
We suggest explanations for these phenomena, but nailing down the fundamental reasons behind them will require further investigations. Such deeper studies will also help understand which properties are critical in the performance and sample efficiency of policy search algorithms, and define even more efficient policy search algorithms in the future. As suggested in Section~\ref{sec:other_res}, another avenue for future work will consist in designing an \evorl algorithm based on \cem rather than on an ad hoc evolutionary algorithm.
Finally, given the impact of the neural architecture on our results, we believe that a more systemic search of architectures through techniques such as neural architecture search \citep{zoph2016neural,elsken2018nassurvey} may provide important progress in performance of deep policy search algorithms.

\section{Acknowledgments}
This work was supported by the European Commission, within the DREAM project, and has received funding from the European Union's Horizon 2020 research and innovation program under grant agreement $N^o$ 640891. We would like to thank Thomas Pierrot for fruitful discussions.

\bibliography{local}
\bibliographystyle{iclr2019_conference}

\appendix

\section{Architecture of the networks}
\label{sec:archi}

Our network architectures are very similar to the ones described in \cite{fujimoto2018adressing}. In particular, the size of the layers remains the same. The only differences resides in the non-linearities. We use $\tanh$ operations in the actor between each layer, where \citeauthor{fujimoto2018adressing} use \relu and we use leaky \relu in the critic, where \citeauthor{fujimoto2018adressing} use simple \relu. Reasons for this choice are presented in Appendix~\ref{sec:relu}.

\begin{table}[hbtp]
\caption{Architecture of the networks (from the input layer (top line) to the output layer (bottom line)\label{tab:archi}}
\begin{center}
\begin{tabular}{cc}
\hline
Actor & Critic \\
\hline
(\text{state dim}, 400) &(\text{state dim + action dim}, 400) \\
$\tanh$ &\text{leaky \relu} \\
(400, 300) &(400, 300) \\
$\tanh$ &\text{leaky \relu} \\
(300, action dim) &(300, 1) \\
$\tanh$ & \\
\hline
\hline
\end{tabular}
\end{center}
\end{table}

\section{Importance mixing}
\label{sec:mixing}

Importance mixing is a specific mechanism designed to improve the sample efficiency of evolution strategies. It was initially introduced in \cite{sun2009efficient} and consisted in 
reusing some samples from the previous generation into the current one, to avoid the cost of re-evaluating the corresponding policies in the environment. The mechanism was recently extended in \cite{pourchot2018importance} to reusing samples from any generation stored into an archive.
Empirical results showed that importance sampling can improve sample efficiency by a factor of ten, and that most of these savings just come from using the samples from the previous generation, as performed by the initial mechanism. A pseudo-code of the importance mixing mechanism is given in Algorithm~\ref{alg:IM}.

\begin{algorithm}[htb]
  \caption{Importance mixing}
  \label{alg:IM}
  \begin{algorithmic}[1]
    \REQUIRE $p(z,\vth_{new})$: current probability density function (pdf), $p(z,\vth_{old})$: old pdf, $g_{old}$: old generation
    \STATE  $g_{new} \leftarrow \emptyset$
    \FOR{$i \leftarrow 1$ to $N$}
    	\item[]
        \STATE Draw rand1 and rand2 uniformly from $[0,1]$
        \item[]
    	\STATE  Let $z_i$ be the $i^{th}$ individual of the old generation $g_{old}$
    	\IF {$min(1,\frac{p(z_i,\vth_{new})} {p(z_i,\vth_{old})})>$ rand1:}
    		\STATE  Append $z_i$ to the current generation $g_{new}$
    	\ENDIF
    	\item[]
    	\STATE  Draw $z'_i$ according to the current pdf $p(.,\vth_{new})$
        \IF {$max(0,1-\frac{p(z'_i,\vth_{old})}{p(z'_i,\vth_{new})})>$ rand2:}  
    		\STATE  Append $z'_i$ to the current generation $g_{new}$
   		\ENDIF
    	\item[]
		\STATE  size = $|g_{new}|$
		\ONELINEIF{size $\geq N$:}{go to 12}
    	\item[]
    \ENDFOR
    \ONELINEIF{size $> N$:}{remove a randomly chosen sample}
    \ONELINEIF{size $< N$:}{fill the generation sampling from $p(.,\vth_{new})$}
    \RETURN  $g_{new}$
  \end{algorithmic}
\end{algorithm}

In \cem, importance mixing is implemented as described in \citep{pourchot2018importance}.
By contrast, some adaptation is required in \cemrl. Actors which take gradient steps can no longer be regarded as sampled from the current distribution of the \cem algorithm. We thus choose to apply importance mixing only to the half of the population which does not receive gradient steps from the \rl critic. In practice, only actors which do not take gradient steps are inserted into the actor archive and can be replaced with samples from previous generations.

\begin{figure}[!ht]
  \centering
  \subfloat[\label{fig:im_hc}]{\includegraphics[width=0.34\linewidth]{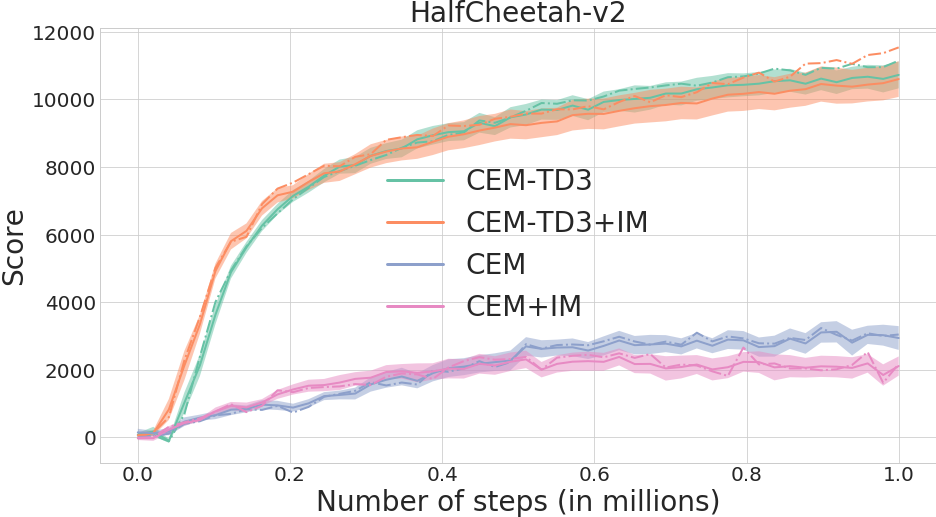}}
  \subfloat[\label{fig:im_hopper}]{\includegraphics[width=0.34\linewidth]{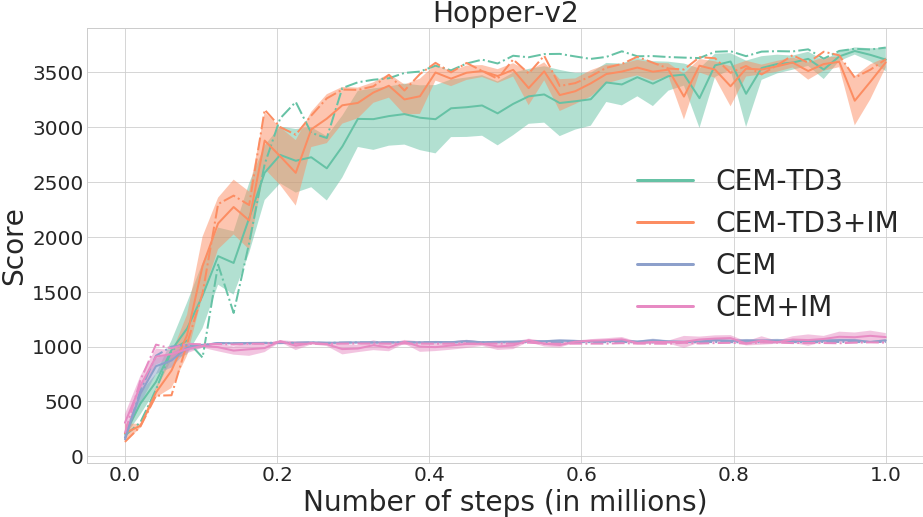}}
   \subfloat[\label{fig:im_walker}]{\includegraphics[width=0.34\linewidth]{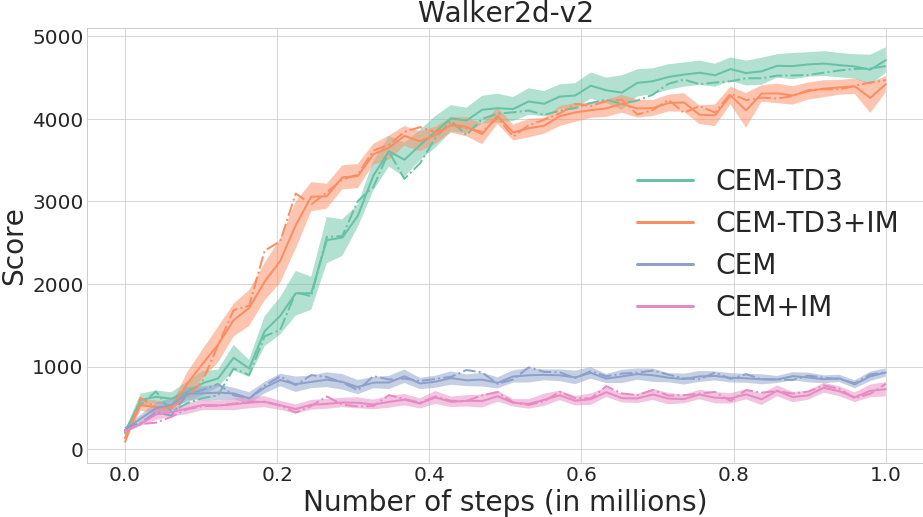}}
   
  \medskip
  
  \subfloat[\label{fig:im_swimmer}]{\includegraphics[width=0.34\linewidth]{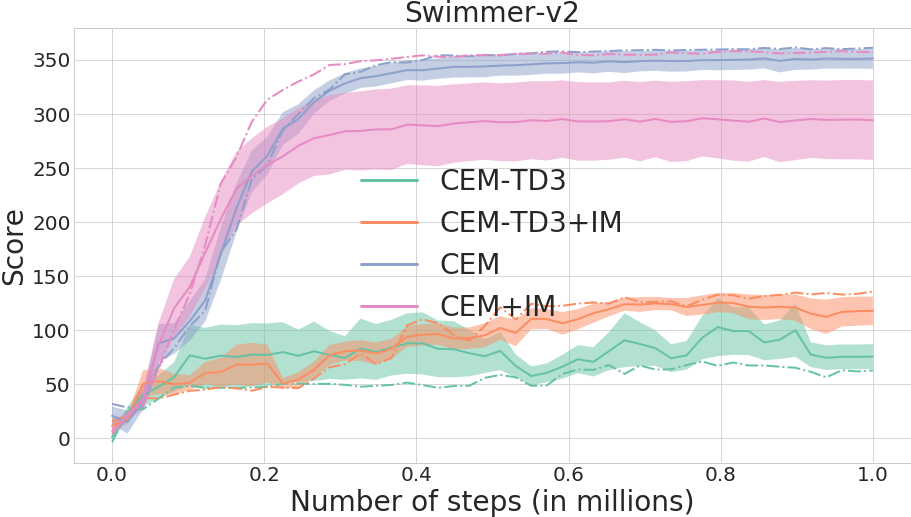}}
    \subfloat[\label{fig:im_ant}]{\includegraphics[width=0.34\linewidth]{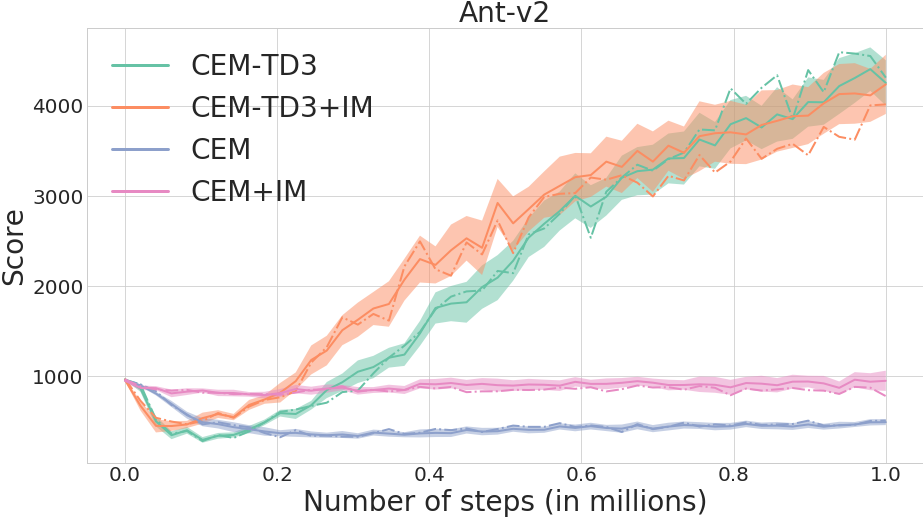}}
  
   \caption{Learning curves of \cemtdd and \cem with and without importance mixing on the \hc, \hop,  \walk, \sw and \ant benchmarks. 
   \label{fig:im}}
\end{figure}

From \figurename~\ref{fig:im}, one can see that in the \cem case, importance mixing introduces some minor instability, without noticeably increasing sample efficiency.  On \hc, \sw and \walk, performance even decreases when using importance mixing. For \cemrl, the effect varies greatly from environment to environment, but the gain in sample reuse is almost null as well, though an increase in performance can be seen on \sw. The latter fact is consistent with the finding that the gradient steps are not useful in this environment (see Appendix~\ref{sec:sw}). On \hop and \hc, results with and without importance mixing seem to be equivalent. On \walk, importance mixing decreases final performance. On \ant, importance mixing seems to accelerate learning in the beginning, but final performances are equivalent to those of \cemrl. Thus importance mixing seems to have a limited impact in \cemtdd.

\begin{table}[ht]
\centering
\begin{tabular}{l r r r r r r }

\hline 
         & \multicolumn{3}{c}{\cemtdd} & \multicolumn{3}{c}{\cemtdd + \sc{im}}  \\
         Environment  & Mean & Var. & Median  & Mean & Var. & Median  \\
         \hline
         \\
  \hc & $\mathbf{10725}$ & $\mathbf{3.7\%}$ & $\mathbf{11147}$ & $10601$ & $4.9\%$ & $11539$ \\
  
  \hop & $\mathbf{3613}$ & $\mathbf{2.9\%}$ & $\mathbf{3722}$ & $3589$ & $1.2\%$ & $3616$ \\
  
  \walk & $\mathbf{4711}$ & $\mathbf{3.3\%}$ & $\mathbf{4637}$ & $4420$ & $2.3\%$ & $\mathbf4468$ \\
  
  \sw & $75$ & $15\%$ & $62$ & $\mathbf{117}$ & $\mathbf{11\%}$ & $\mathbf{135}$ \\
  
  \ant & $\mathbf{4251}$ & $\mathbf{5.9\%}$ & $\mathbf{4310}$ & $4235$ & $7.8\%$ & $4013$  \\
  
  \\
  \hline
\hline 
  \end{tabular}
\caption{\label{table:mixing} Final performance of \cemtdd with and without importance mixing on the \hc, \hop, \sw, \ant and \walk environments. We report the mean ands medians over 10 runs of 1 million steps. 
For each benchmark, we highlight the results of the method with the best mean.}
\end{table}

This conclusion seems to contradict the results obtained in \cite{pourchot2018importance}. This may be due to different things. First, the dimensions of the search spaces in the experiments here are much larger than those studied in \cite{pourchot2018importance}, which might deteriorate the estimation of the covariance matrices when samples are too correlated. On top of this, the \muj environments are harder than the ones used in \cite{pourchot2018importance}. In particular, we can see from \figurename~\ref{fig:cem_tdd_cmerl} 
that \cem is far from solving the environments over one million steps. Perhaps a study over a longer time period would make importance mixing relevant again. Besides, by reusing old samples, the importance mixing mechanism somehow hinders exploration (since we evaluate less new individuals), which might be detrimental in the case of \muj environments.
Finally, and most importantly, the use of \rl gradient steps accelerates the displacement of the covariance matrix, resulting in fewer opportunities for sample reuse.

\section{Effect of action noise}
\label{sec:noise}

\begin{figure}[!ht]
  \centering
  \subfloat[\label{fig:a_noise_hc}]{\includegraphics[width=0.34\linewidth]{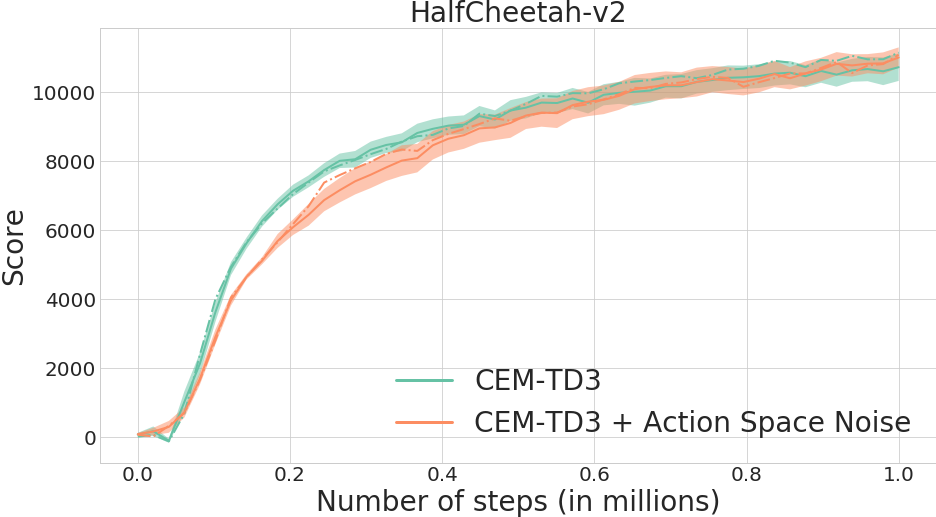}}
  \subfloat[\label{fig:a_noise_hopper}]{\includegraphics[width=0.34\linewidth]{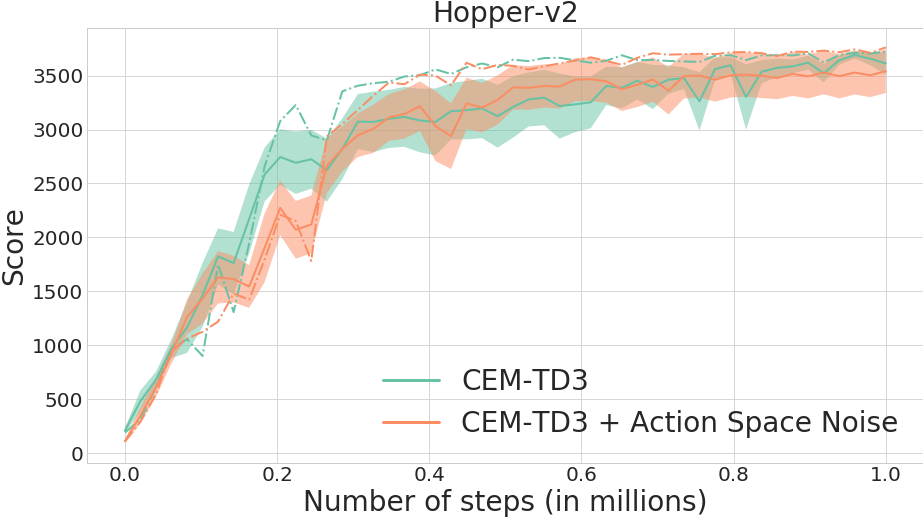}}
   \subfloat[\label{fig:a_noise_walker}]{\includegraphics[width=0.34\linewidth]{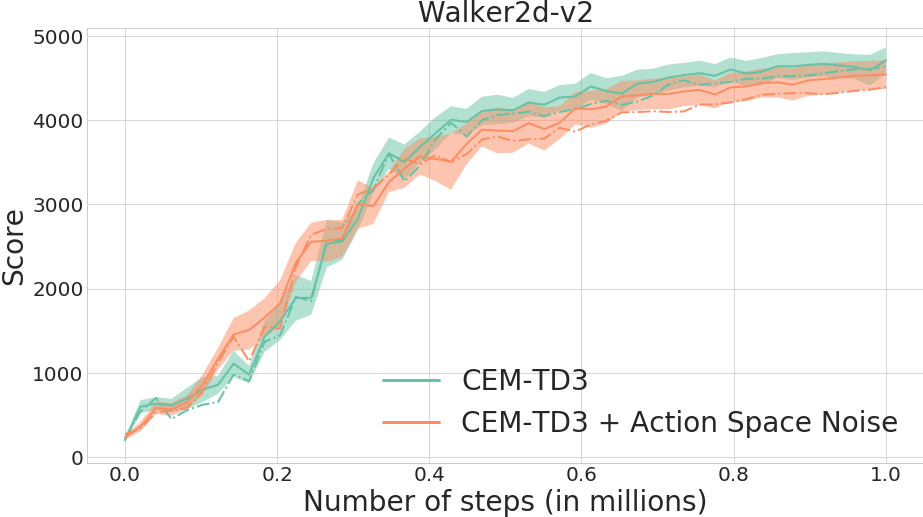}}
  
  \medskip
  
  \subfloat[\label{fig:a_noise_swimmer}]{\includegraphics[width=0.34\linewidth]{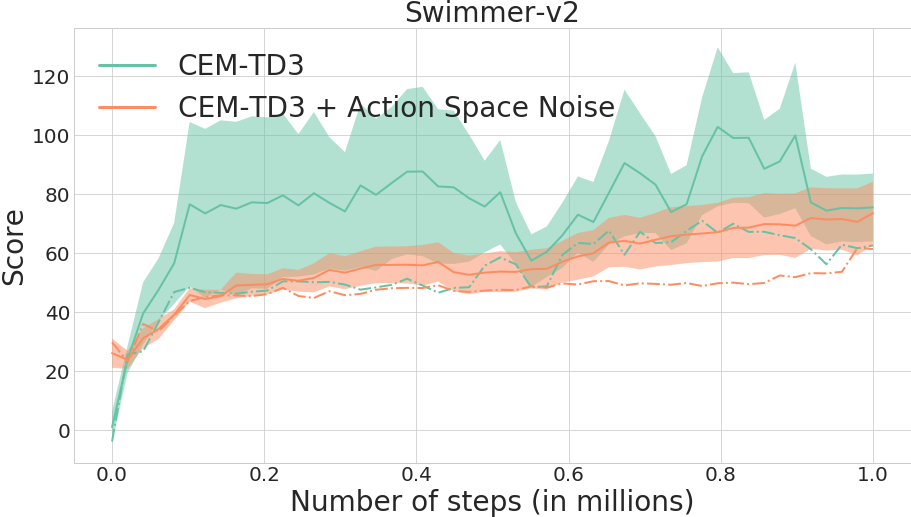}}
    \subfloat[\label{fig:a_noise_ant}]{\includegraphics[width=0.34\linewidth]{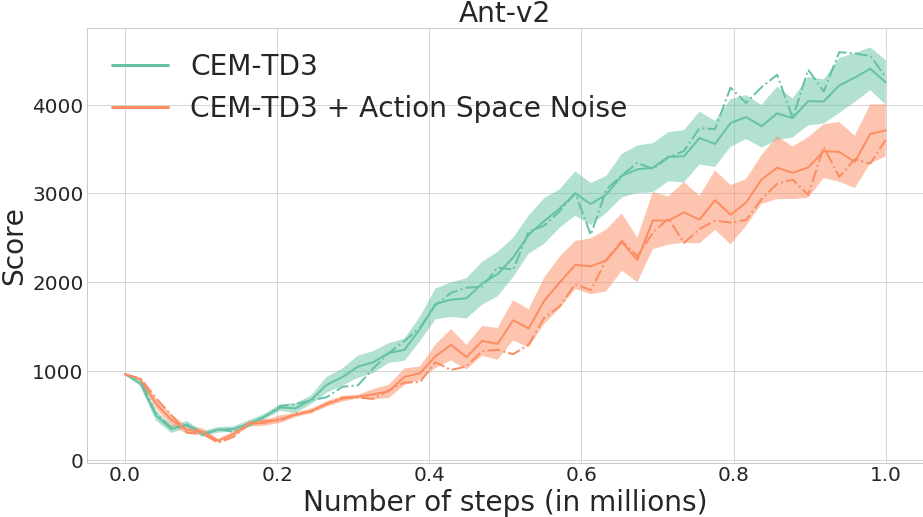}}
 
   \caption{Learning curves of \cemrl with and without action space noise on the \hc, \hop,  \walk, \sw and \ant benchmarks. 
   \label{fig:a_noise}}
\end{figure}

In \cite{khadka2018evolutionaryNIPS}, the authors indicate that one reason for the efficiency of their approach is that the replay buffer of \ddpg gets filled with two types of noisy experiences. On one hand, the buffer gets filled with noisy interactions of the \ddpg actor with the environment. This is usually referred to as {\em action space noise}. On the other hand, actors with different parameters also fill the buffer, which is more similar to {\em parameter space noise} \citep{plappert2017parameter}. In \cemrl, we only use parameter space noise, but it would also be possible to add action space noise. To explore this direction, each actor taking gradient steps performs a noisy episode in the environment. We report final results after 1 million steps in Table~\ref{table:other_factors}. Learning curves are available in \figurename~\ref{fig:a_noise}.

Unlike what \cite{khadka2018evolutionaryNIPS} suggested, we did not find any conclusive evidence that action space noise improves performance. In \cemtdd, the \cem part seems to explore enough of the action space on its own. It seems that sampling performed in \cem results in sufficient exploration and performs better than adding simple Gaussian noise to the actions. This highlights a difference between using an evolutionary strategy like \cem and an evolutionary algorithm as done in \evorl. Evolutionary algorithms tend to converge to a unique individual whereas evolutionary strategies more easily maintain some exploration. These aspects are further studied in Appendix~\ref{sec:dynam}.

\begin{table}[!ht]
{\small
\centering
\scalebox{0.95}{%
\begin{tabular}{l  r r r  r r r  r r r}

\hline 
         & \multicolumn{3}{c}{\cemtdd} & \multicolumn{3}{c}{\cemtdd + \sc{an}} & \multicolumn{3}{c}{\cemtdd - \relu} \\
         Environment & Mean & Var. & Median  & Mean & Var. & Median & Mean & Var. & Median \\
         \hline
         &&&&&&&&&\\
  \hc & $\mathbf{10725}$ & $\mathbf{3.7\%}$ & $\mathbf{11147}$ & $11006$ & $2.7\%$ & $11086$ & $10267$ & $3.7\%$ & $10133$ \\
  
  \hop & $\mathbf{3613}$ & $\mathbf{2.9}$ \% & $\mathbf{3722}$ & $3541$ & $5.7\%$ & $3761$ & $3604$ & $2.6\%$ & $3716$ \\
  
  \walk & $\mathbf{4711}$ & $\mathbf{3.3\%}$ & $\mathbf{4637}$ & $4542$ & $5.7\%$ & $4392$ & $4311$ & $7.5\%$ & $4534$ \\
  
  \sw & $75$ & $15\%$ & $62$ & $74$ & $15\%$ & $62$ & $\mathbf{118}$ & $\mathbf{21\%}$ & $\mathbf{114}$ \\
  
  \ant & $\mathbf{4251}$ & $\mathbf{5.9\%}$ & $\mathbf{4310}$ & $3711$ & $7.9\%$ & $3604$ & $2264$ & $16\%$ & $2499$ \\
  &&&&&&&&&\\
  \hline
  \hline
  \end{tabular}
  }
\caption{\label{table:other_factors}
Final Performance of \cemrl with and without action noise ({\sc{an}}), with \ddpg, and with \relu non-linearities in \muj environments. We report the mean ands medians over 10 runs of 1 million steps. 
For each benchmark, we highlight the results of the method with the best mean.}
}
\end{table}

\section{Parameter space exploration in \cemrl and \evorl}
\label{sec:dynam}

\begin{figure}[!ht]
  \centering
  \subfloat[\label{fig:erl_path}]{\includegraphics[width=0.5\linewidth]{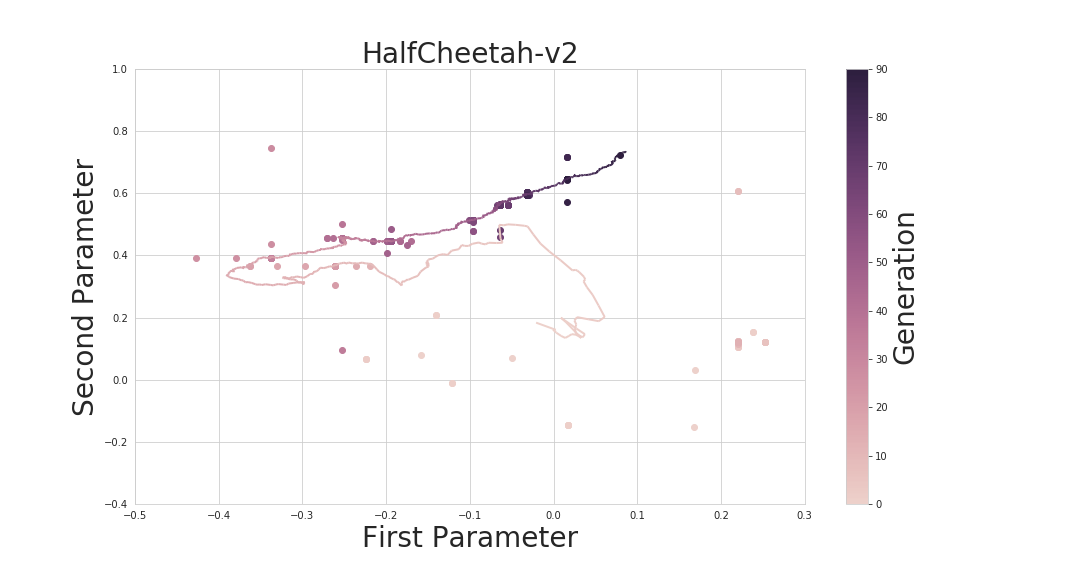}}
  \subfloat[\label{fig:cemrl_path}]{\includegraphics[width=0.5\linewidth]{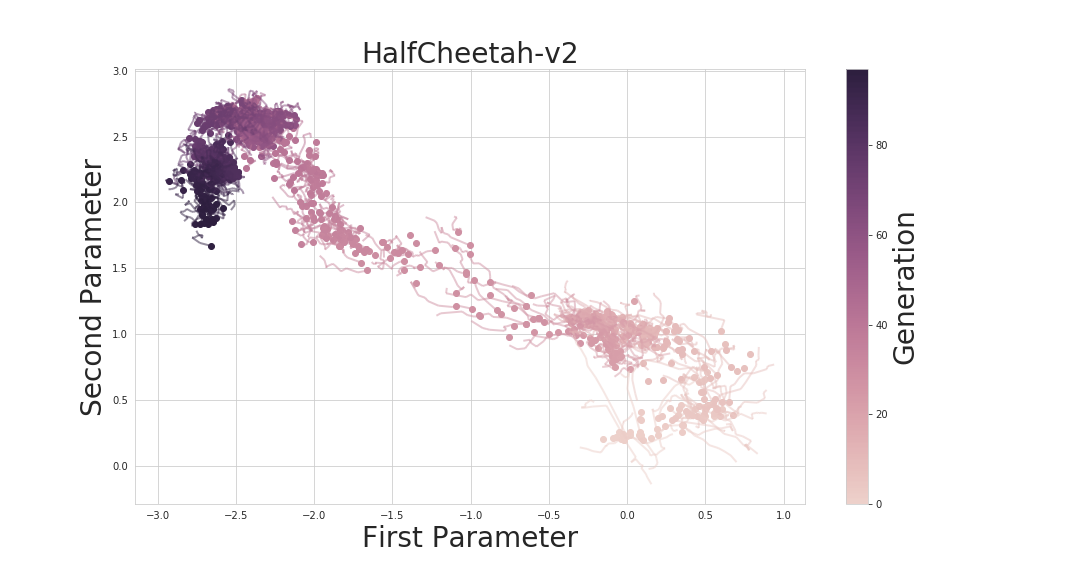}}
   \caption{Evolution of the first two parameters of the actors when learning with (a) \evorl and (b) \cemtdd. Dots are sampled parameters of the population and continuous lines represent parameters moved through \rl gradient steps.\label{fig:dynam}}
\end{figure}

In this section, we highlight the difference in policy parameter update dynamics in \cemrl and \evorl. \figurename~\ref{fig:dynam} displays the evolution of the first two parameters of actor networks during training with \cemrl and \evorl on \hc. For \evorl, we plot the chosen parameters of the \ddpg actor with a continuous line, and represent those of the evolutionary actors with dots. For \cemrl, we represent the chosen parameters of sampled actors with dots, and the gradient steps based on the \tdd critic with continuous lines. The same number of dots is represented for both algorithms.

One can see that, in \evorl the evolutionary population tends to be much less diverse that in \cemrl. There are many redundancies in the parameters (dots with the same coordinates), and the population seems to converge to a single individual. On the other hand, there is no such behavior in \cemrl where each generation introduces completely new samples. As a consequence, parameter space exploration looks better in the \cemrl algorithm.

\begin{figure}[!ht]
  \centering
 \includegraphics[width=0.7\linewidth]{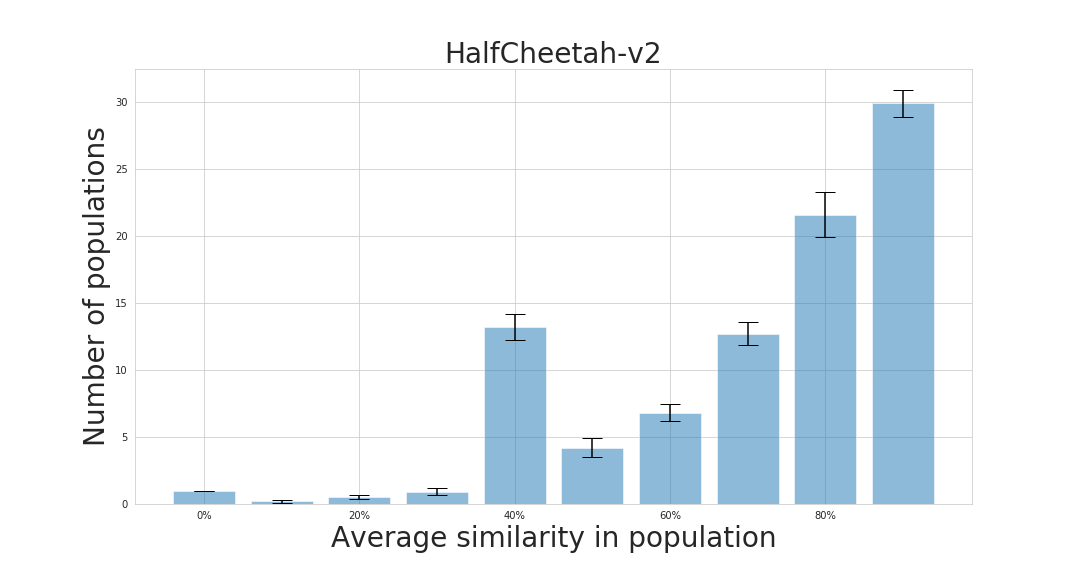}
   \caption{Histogram of the average similarity in populations during learning with the \evorl algorithm. Results are averaged over 10 runs. As usual, the variance corresponds to the $68\%$ confidence interval for the estimation of the mean.\label{fig:erl_hist}}
\end{figure}

To further study this loss of intra-population diversity in \evorl, we perform 10 \evorl runs and report in \figurename~\ref{fig:erl_hist} an histogram displaying the distribution of the population-wise similarity with respect to the populations encountered during learning. We measure this similarity as the average percentage of parameters shared between two different individuals of the said population. The results indicate that around $55\%$ of populations encountered during a run of \evorl display a population-similarity of above $80\%$.

One can also see the difference in how both methods use the gradient information of their respective \drl part. In the case of \evorl, the parameters of the population concentrate around those of the \ddpg actor. Each 10 generations, its parameters are introduced into the population, and since \ddpg is already efficient alone on \hc, those parameters quickly spread into the population. Indeed, according to \cite{khadka2018evolutionaryNIPS}, the resulting \ddpg actor is the elite of the population $80\%$ of the time, and is introduced into the population $98\%$ of the time. This integration is however passive: the direction of exploration does not vary much after introducing the \ddpg agent. \cemrl integrates this gradient information differently. The short lines emerging from dots, which represent gradient steps performed by half of the actors, act as scouts. Once \cem becomes aware of better solutions that can be found in a given direction, the sampling of the next population is modified so as to favor this promising direction. \cem is thus pro-actively exploring in the good directions it has been fed with.

\section{The case of the \sw benchmark}
\label{sec:sw}

\begin{figure}[!ht]
  \centering
 \subfloat[\label{fig:swimmer_comparison}]{\includegraphics[width=0.48\linewidth]{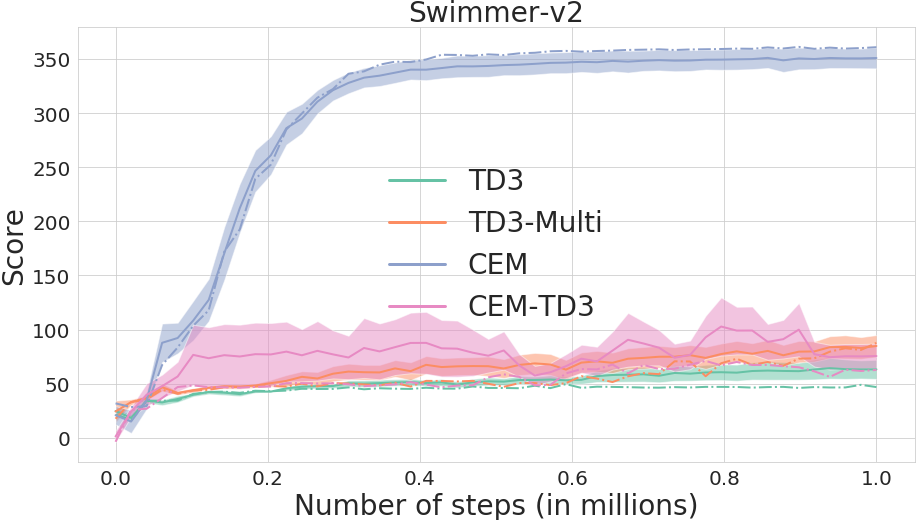}}
\subfloat[\label{fig:erl_cemrl_swimmer}]{\includegraphics[width=0.48\linewidth]{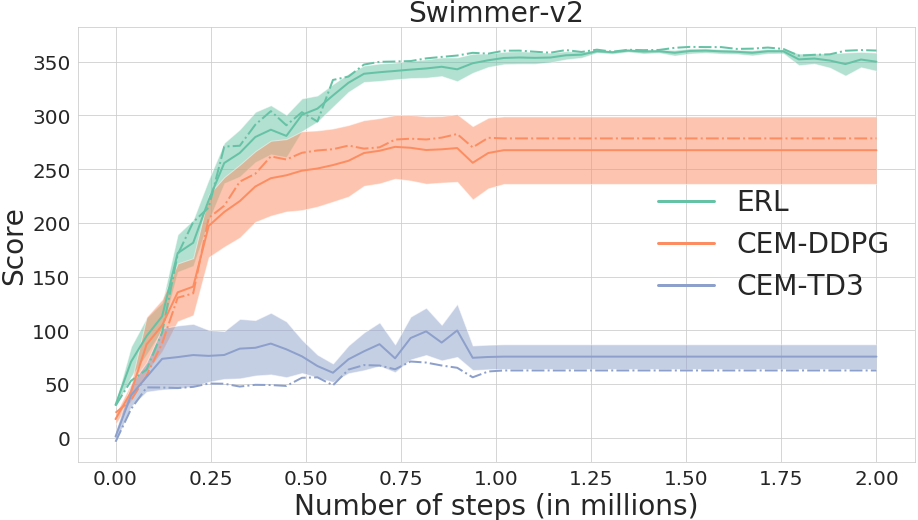}}
\caption{Learning curves on the \sw environment of (a): \cem and \tdd, multi-actor \tdd and \cemrl; (b) \evorl, \cemddpg and \cemtdd. 
\label{fig:sw}}
\end{figure}

Experiments on the \sw benchmark give results that differ a lot from the results on other benchmarks, hence they are covered separately here. \figurename\ref{fig:swimmer_comparison} shows that \cem outperforms \tdd, \cemtdd, multi-actor \tdd. Besides, as shown in \figurename~\ref{fig:erl_cemrl_swimmer}, \evorl outperforms \cemddpg, which itself outperforms \cemtdd.

All these findings seem to show that being better at RL makes you worse at \sw. The most likely explanation is that, in \sw, any deep RL method provides a deceptive gradient information which is detrimental to convergence towards efficient actor parameters. This conclusion could already be established from the results of \cite{khadka2018evolutionaryNIPS}, where the evolution algorithm alone produced results on par with the \evorl algorithm, showing that RL-based actors were just ignored.
In this particular context, the actors using \tdd gradient being deteriorated by the deceptive gradient effect, \cemrl is behaving as a \cem with only half a population, thus it is less efficient than the standard \cem algorithm. By contrast, \evorl better resists than \cemrl to the same issue. Indeed, if the actor generated by \ddpg does not perform better than the evolutionary population, then this actor is just ignored, and the evolutionary part behaves as usual, without any loss in performance. In practice, \citeauthor{khadka2018evolutionary} note that on \sw, the \ddpg actor was rejected $76\%$ of the time.
Finally, by comparing \cem and \evorl from \figurename~\ref{fig:swimmer_comparison} and \figurename~\ref{fig:erl_cemrl_swimmer}, one can conclude that on this benchmark, the evolutionary part of \evorl behaves on par with \cem alone. This is at odds with premature convergence effects seen in the evolutionary part of \evorl, as studied in more details in Appendix~\ref{sec:dynam}.
From all these insights, the \sw environment appears particularly interesting, as we are not aware of any deep \rl method capable of solving it quickly and reliably.

\section{Using the \relu or tanh non-linearity}
\label{sec:relu}

In this section, we explore the impact on performance of the type of non-linearities used in the actor of \cemtdd. Table~\ref{table:other_factors} reports the results of \cemtdd using \relu non-linearities between the linear layers, instead of $\tanh$. 

\begin{figure}[!ht]
  \centering
  \subfloat[\label{fig:tanh_relu_hc}]{\includegraphics[width=0.34\linewidth]{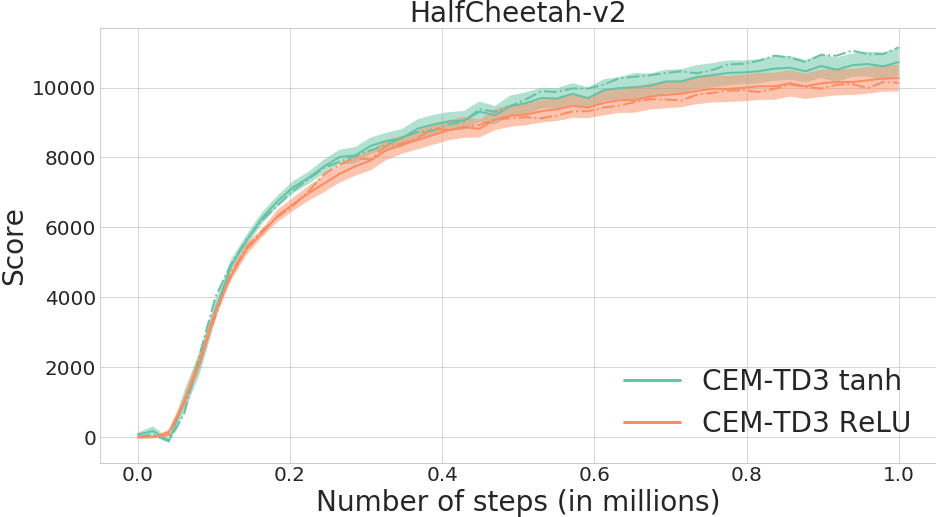}}
  \subfloat[\label{fig:tanh_relu_hopper}]{\includegraphics[width=0.34\linewidth]{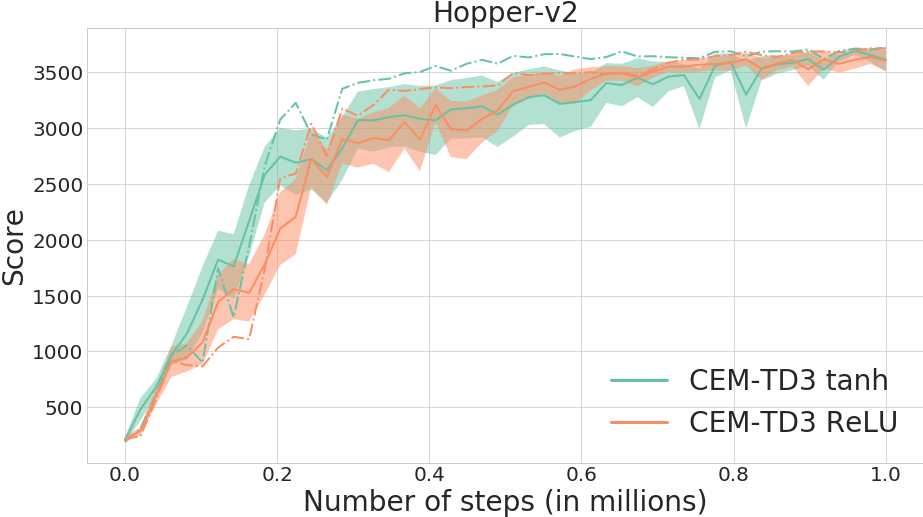}}
  \subfloat[\label{fig:tanh_relu_swimmer}]{\includegraphics[width=0.34\linewidth]{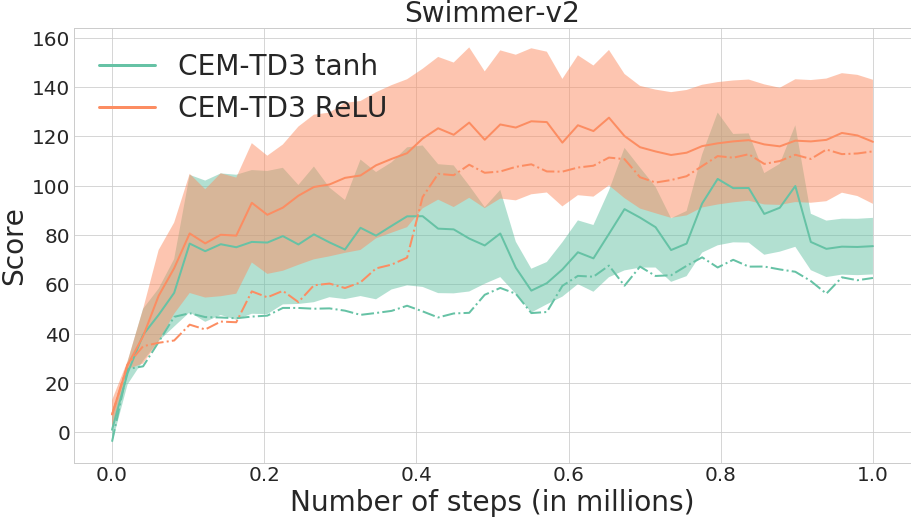}}
  
  \medskip
    \subfloat[\label{fig:tanh_relu_ant}]{\includegraphics[width=0.34\linewidth]{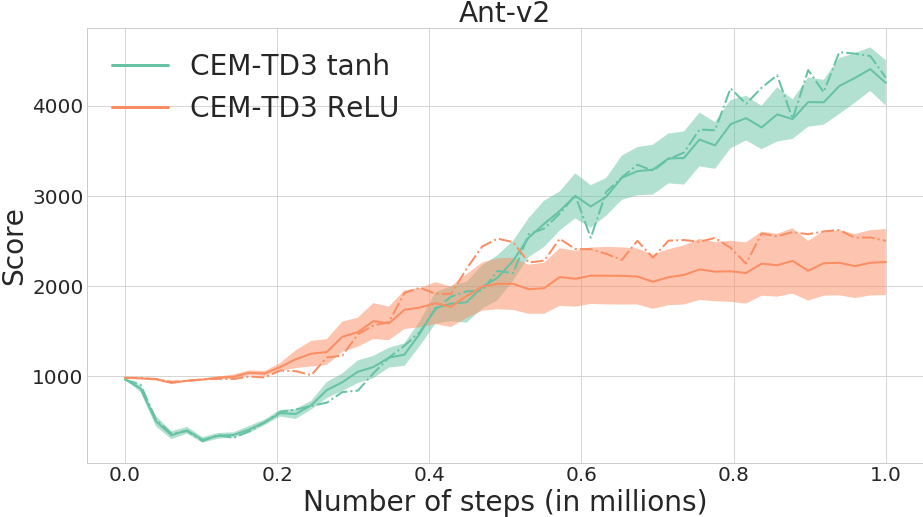}}
  \subfloat[\label{fig:tanh_relu_walker}]{\includegraphics[width=0.34\linewidth]{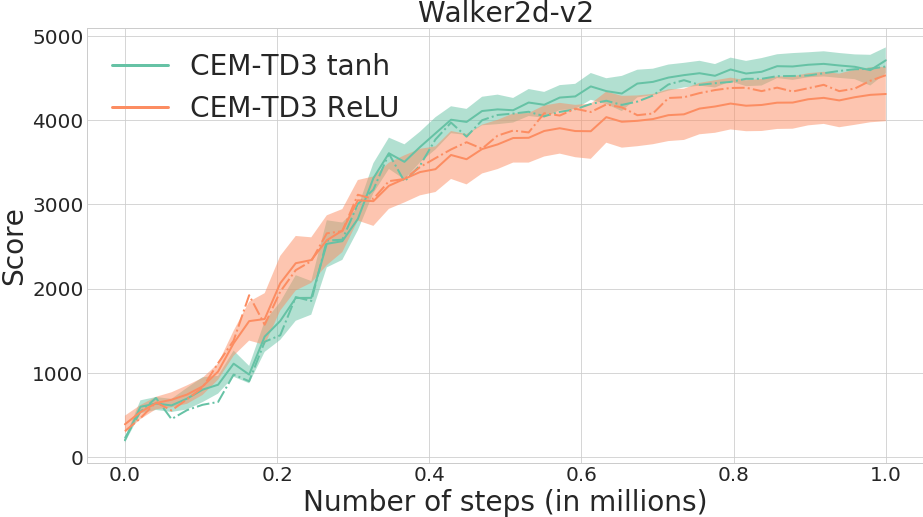}}
    \subfloat[\label{fig:tanh_relu_cem_swimmer}]{\includegraphics[width=0.34\linewidth]{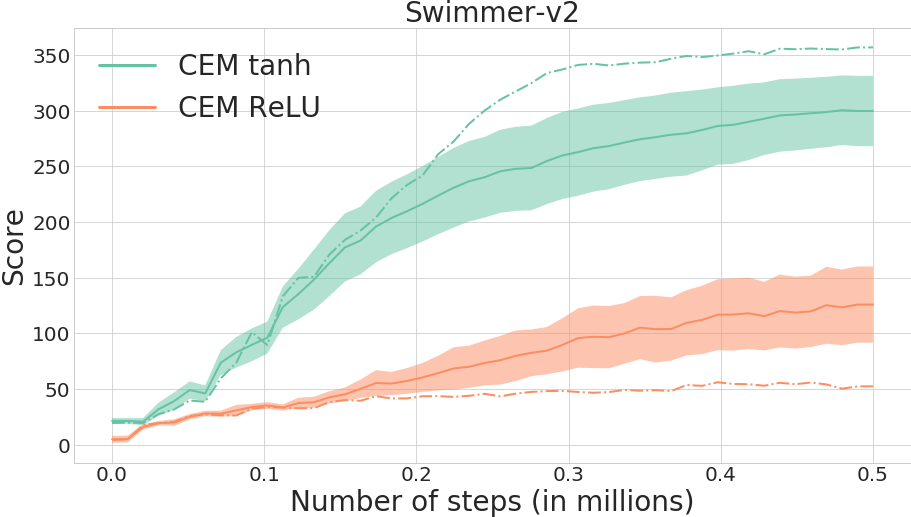}}
   \caption{Learning curves of \cemrl with $\tanh$ and \relu as non-linearities in the actors, on the (a) \hc, (b) \hop, (c) \sw, (d) \ant and (e) \walk benchmarks. (f) shows the same of \cem on the \sw benchmark. 
   \label{fig:tanh_relu}}
\end{figure}

\figurename~\ref{fig:tanh_relu} displays the learning performance of \cemtdd and \cem on benchmarks, using either the \relu or the $tanh$ nonlinearity in the actors. 
Results indicate that on some benchmarks, changing from $\tanh$ to \relu can cause a huge drop in performance. This is particularly obvious in the \ant benchmark, where the average performance drops by 46$\%$. \figurename~\ref{fig:tanh_relu}(f) shows that, for the \cem algorithm on the \sw benchmark, using \relu also causes a 60$\%$ performance drop. As previously reported in the literature \citep{henderson2017deep}, this study suggests that network architectures can have a large impact on performance. 

\section{Additional results on \ant}
\label{app:ant}

\begin{figure}[!ht]
  \centering
\subfloat[\label{fig:ant_comparison}]{\includegraphics[width=0.48\linewidth]{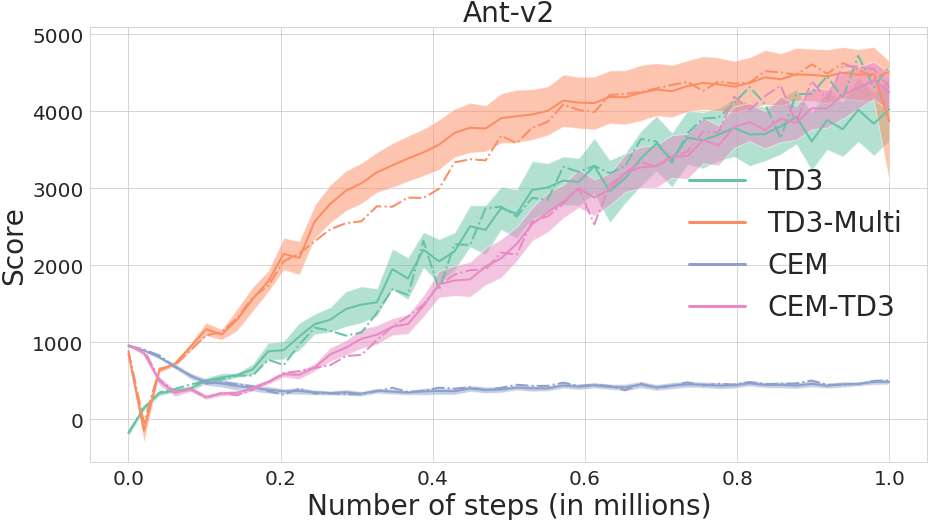}}
\subfloat[\label{fig:erl_cemrl_ant}]{\includegraphics[width=0.48\linewidth]{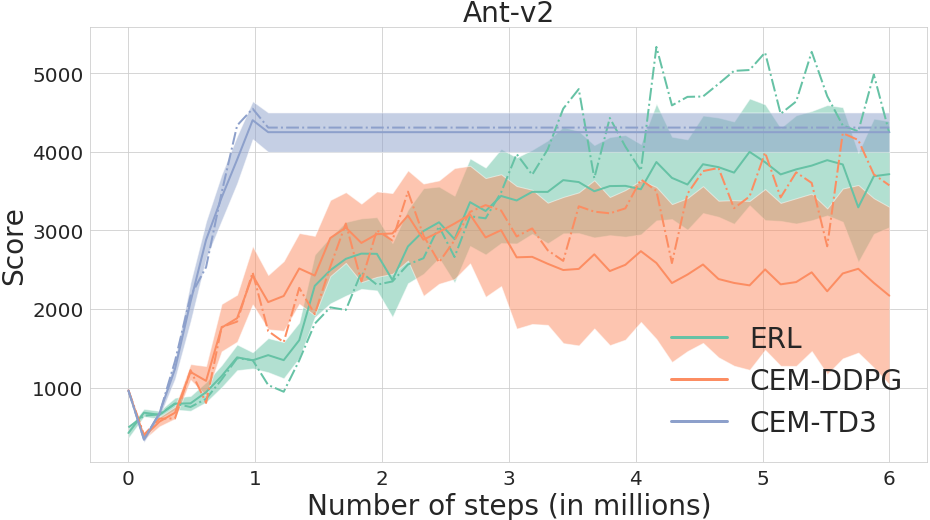}}
\caption{Learning curves of \cemrl, \cem and \tdd on the \sw and \ant benchmarks. \label{fig:ant}}
\end{figure}

\figurename~\ref{fig:ant} represents the learning performance of \cem, \tdd, multi-actor \tdd, \cemddpg and \cemtdd on the \ant benchmark. It is discussed in the main text.

\end{document}